\documentclass[journal]{IEEEtran}

\usepackage{amsmath,amsfonts,amssymb}
\usepackage{bm}

\usepackage{graphicx}
\usepackage{subcaption}        

\usepackage{booktabs}          
\usepackage{multirow}
\usepackage{array}
\usepackage[table]{xcolor}     

\usepackage{cite}              
\usepackage{url}
\definecolor{best}{HTML}{FFCCCC}     
\definecolor{second}{HTML}{FFE5CC}   
\definecolor{third}{HTML}{FFFFCC}    
\definecolor{rowgray}{HTML}{F5F5F5}  

\begin{document}

\title{GrainGS: Gradient-Decoupled Gaussian Splatting
       for Efficient Dynamic Novel View Synthesis}

\author{Jiahao~He,~Yihua Shao,~Zhengkai~Zhao,~Pan~Gao,~Fei Ma, Jingcai Guo,~\IEEEmembership{Senior Member,~IEEE},\\Hao Tang,~\IEEEmembership{Senior Member,~IEEE}, Nicu Sebe,~\IEEEmembership{Senior Member,~IEEE}, Qi Tian,~\IEEEmembership{Fellow,~IEEE}
\thanks{This work was supported in part by the National Natural Science Foundation of China (No. 62272227) and supported in part by the Guangdong Basic and Applied Basic Research Foundation (2026A1515010184) and the Research Task Assignment Project from Guangdong Laboratory of Artificial Intelligence and Digital Economy (SZ) (GML-26420007).}
\thanks{Jiahao He, Zhengkai Zhao, and Pan Gao are with the College of Artificial Intelligence,
        Nanjing University of Aeronautics and Astronautics, Nanjing, Jiangsu, China
        (e-mail: nuaahjh@nuaa.edu.cn; zhengkai.zhao@nuaa.edu.cn; pan.gao@nuaa.edu.cn).}
\thanks{Yihua Shao is with Institute of Automation, Chinese Academy of Sciences, Beijing, China, and also with Department of Computing, Hong Kong Polytechnic University, Hong Kong, China. (e-mail: yihuajerry@gmail.com)}
\thanks{Hao Tang is with the School of Computer Science, Peking University, Beijing, China. (e-mail: bjdxtanghao@gmail.com)}
\thanks{Fei Ma and Qi Tian are with Guangdong Laboratory of Artificial Intelligence and Digital Economy, Shenzhen, China. Qi Tian is also with Huawei Consumer Business Group, Huawei, Shenzhen, China. (e-mail: mafei@gml.ac.cn, wywqtian@gmail.com)}
\thanks{Jingcai Guo is with Department of Computing, Hong Kong Polytechnic University, Hong Kong, China.  (e-mail: jingcai.guo@gmail.com)}
\thanks{Nicu Sebe is with Department of Information Engineering and Computer Science, University of Trento, Trento, Italy. (e-mail: sebe@disi.unitn.it)}
\thanks{Jiahao He and Yihua Shao are Equal Contribution (Corresponding authors: Pan Gao and Fei Ma.)}}

\markboth{}%
{He \MakeLowercase{\textit{et al.}}: GrainGS: Gradient-Decoupled Gaussian Splatting}

\maketitle

\begin{abstract}

Dynamic scene reconstruction with 3D Gaussian
Splatting requires a balance between fine-grained motion modeling,
structural stability, and compact representation. Existing
per-primitive methods provide flexible local deformation but often
suffer from redundant primitive growth, while anchor-based methods
improve spatial regularity at the cost of suppressing locally varying
motion. To address these issues, we present GrainGS, a dynamic Gaussian framework that
combines a hierarchical anchor scaffold with per-Gaussian deformation.
A static warm-up stage first establishes a time-invariant canonical
representation from observations across all timestamps. During joint
training, a stop-gradient operation blocks the deformation-mediated
gradient pathway to the canonical positions while preserving their
direct refinement through the reconstruction objective. Each Gaussian
then predicts independent temporal offsets for position, rotation, and
scale, enabling detailed local motion within a structurally constrained
scaffold. A canonical-residual appearance decomposition further models
frame-dependent photometric changes without forcing them into geometric
deformation. Experiments on synthetic monocular and real-world
multiview benchmarks show that GrainGS achieves high reconstruction
quality, real-time novel view synthesis, and compact storage. Under the
synthetic benchmark setting, it reaches an average peak
signal-to-noise ratio of 36.98 decibels, renders at 435.6 frames per
second, and requires 4.67 megabytes of storage.
\end{abstract}

\begin{IEEEkeywords}
3D Gaussian Splatting, 4D Reconstruction, Neural Rendering.
\end{IEEEkeywords}

\section{Introduction}
\label{sec:intro}

\IEEEPARstart{C}{apturing} dynamic real-world scenes and rendering them from
arbitrary viewpoints are essential for free-viewpoint video and visual
effects~\cite{li2022neuralthreedvideos,guo2024realtimefvv,
liu2025humanradianceocclusion,jiang2024gssfs,shin2025neuralvolumetricvideo,
peng2021neuralbody}.
With multi-view capture systems becoming increasingly accessible,
there is growing demand for methods that reconstruct high-fidelity
geometry, motion, and appearance while supporting real-time
rendering~\cite{li2022neuralthreedvideos,xu20244kfourd,
liu2025dgmesh,gao2022dycheck}.
Neural Radiance Fields
(NeRFs)~\cite{mildenhall2020nerf} provide a principled framework
for novel-view synthesis~\cite{mildenhall2019llff,fu2024cbarf,
zhao2026msasplatting,verbin2022refnerf,huang2026structgs},
but their volumetric rendering cost limits real-time
applications~\cite{park2021nerfies,li2021nsff,park2021hypernerf,
chen2025atmnerf,liu2023rodynrf,song2023nerfplayer}.
3D Gaussian Splatting
(3D-GS)~\cite{kerbl2023threedgaussianspla} replaces implicit
volumes with explicit Gaussian primitives rendered through
differentiable rasterization~\cite{lassner2021pulsar,
zwicker2001surfacesplatting,zwicker2002ewasplatting,
ren2002objectspaceewasurf}, enabling efficient rendering and
motivating a growing body of dynamic
extensions~\cite{luiten2024dynamicthreedgauss,lin2024gaussianflow,
yang2024deformablethreedga,wu2024fourdgaussiansplat,
huang2024scgs,yan2025streetgaussians,cho2026fourdscaffoldgs,
Kwak_2025_CVPR}.

\begin{figure}[t]
  \centering
  \includegraphics[width=0.95\linewidth]{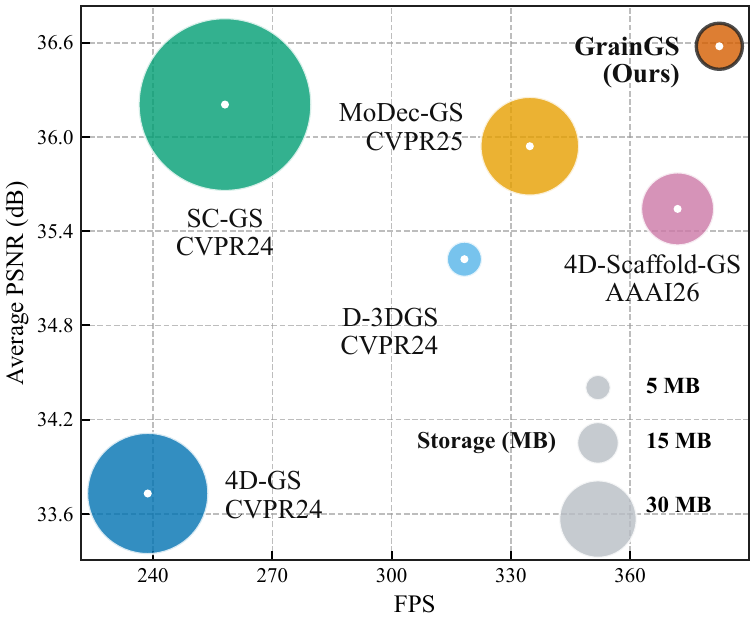}
  \caption{\textbf{Quality-Efficiency Comparison.}
  Average PSNR versus rendering speed for methods with complete statistics;
  bubble size indicates model storage. Values are scene-count-weighted averages
  over eight D-NeRF and six DG-Mesh scenes, reflecting overall performance.}
  \label{fig:fig1}
\end{figure}

Existing dynamic 3D-GS methods must balance motion expressiveness
and structural stability.
Per-primitive approaches provide independent motion degrees of
freedom~\cite{luiten2024dynamicthreedgauss,lin2024gaussianflow,
yang2024deformablethreedga,wu2024fourdgaussiansplat,
yan2025streetgaussians}, but weak structural constraints may lead
to redundant Gaussian growth and unstable canonical geometry.
In contrast, anchor-based methods impose spatial regularity through
sparse control nodes or hierarchical
scaffolds~\cite{huang2024scgs,lu2024scaffoldgs,ren2025octreegs},
yet commonly propagate similar deformation signals to neighboring
Gaussians, limiting fine-grained local motion.
Recent methods such as 4D Scaffold-GS~\cite{cho2026fourdscaffoldgs}
and MoDec-GS~\cite{Kwak_2025_CVPR} improve compactness through
dynamic anchor allocation or motion decomposition, but canonical
geometry and temporal deformation remain closely coupled during
optimization.

We identify \textbf{deformation-mediated gradient interference} as
an important source when canonical coordinates
are differentiable inputs to the deformation network, frame-specific
deformation gradients can perturb the canonical representation and
weaken its role as a consistent geometric reference.
Moreover, temporal appearance changes, such as moving shadows and
specular highlights, may be incorrectly absorbed by geometric or
deformation parameters when no dedicated temporal appearance models
are available~\cite{gao2022dycheck,liu2023rodynrf,liu2025dgmesh,
yan2025streetgaussians}.

To address these issues, we propose \textbf{GrainGS}, which combines
a structurally regular anchor scaffold with fine-grained
per-Gaussian deformation.
First, a static warm-up phase optimizes the canonical modules over
images sampled from all timestamps while deformation is disabled,
establishing a time-invariant scaffold before temporal modeling
begins.
During joint training, a stop-gradient boundary blocks the indirect
gradient pathway from the deformation network to the canonical
positions, while retaining their direct refinement through the
reconstruction loss.
Second, each child Gaussian predicts independent temporal offsets
$(\Delta\mathbf{x},\Delta\mathbf{r},\Delta\boldsymbol{\rho})$ from its own
canonical position, allowing locally varying motion within the
structural constraints of the anchor scaffold.
Finally, we decompose color and the neural opacity score into
time-invariant canonical components
$(\mathbf{c}_{\mathrm{can}},o_{\mathrm{can}})$ and
regularized temporal residuals
$(\Delta\mathbf{c},\Delta o)$.
The residual branch provides explicit capacity for frame-dependent
photometric changes, reducing the pressure to explain such variation
through geometric deformation.

We evaluate our methods on different datasets, results show that  GrainGS achieves a strong balance between reconstruction quality and efficiency. In conclusion, our contributions can be summarized as follows:

\begin{enumerate}

\item We propose \textbf{GrainGS}, a per-Gaussian deformation method within a hierarchical anchor scaffold, which preserves structural compactness while capturing fine-grained local motion. In addition, we introduce a canonical-residual appearance decomposition to model temporal photometric variations. 

\item We introduce a two-stage optimization strategy that combines static warm-up with stop-gradient isolation, thereby reducing deformation-mediated interference with the canonical representation. 

\item Extensive experiments on D-NeRF~\cite{pumarola2021dnerf} and DG-Mesh~\cite{liu2025dgmesh} datasets demonstrate that our method achieves competitive state-of-the-art reconstruction quality, real-time rendering performance, and compact model storage. \end{enumerate}

\section{Related Work}

\subsection{Neural Scene Representations}
NeRF~\cite{mildenhall2020nerf} optimizes continuous volumetric
functions via differentiable ray marching, establishing a
principled framework for photorealistic novel view synthesis~\cite{mildenhall2019llff,fu2024cbarf,zhao2026msasplatting,verbin2022refnerf}.
Subsequent work improved training efficiency through
multi-resolution hash encodings~\cite{muller2022instantngp} and
extended the framework to dynamic scenes via time-conditioned
deformation fields~\cite{pumarola2021dnerf,fang2022tineuvox,park2021nerfies,li2021nsff} and
higher-dimensional canonical coordinates~\cite{park2021hypernerf,cao2023hexplane,fridovichkeil2023kplanes},
at the cost of dense volumetric sampling that precludes
real-time rendering~\cite{chen2025atmnerf,liu2023rodynrf,song2023nerfplayer}. 3D-GS~\cite{kerbl2023threedgaussianspla} resolves
this by representing scenes as anisotropic Gaussian primitives
rendered via tile-based differentiable rasterization, achieving
real-time performance while preserving competitive visual
fidelity. Scaffold-GS~\cite{lu2024scaffoldgs} introduced a
hierarchical anchor structure in which neural Gaussians are
spawned from sparse anchor points via learned offset MLPs,
reducing redundant primitive proliferation and imposing
structural regularity~\cite{ren2025octreegs}. This representation targets static
scenes, extending it to dynamic scenes requires introducing
temporal deformation while preserving the geometric stability
the anchor structure provides~\cite{ren2025octreegs,yang2024deformablethreedga,wu2024fourdgaussiansplat,huang2024scgs,cho2026fourdscaffoldgs,yang2024gs4d}.

\begin{figure}[t]
  \centering
  \includegraphics[width=\linewidth]{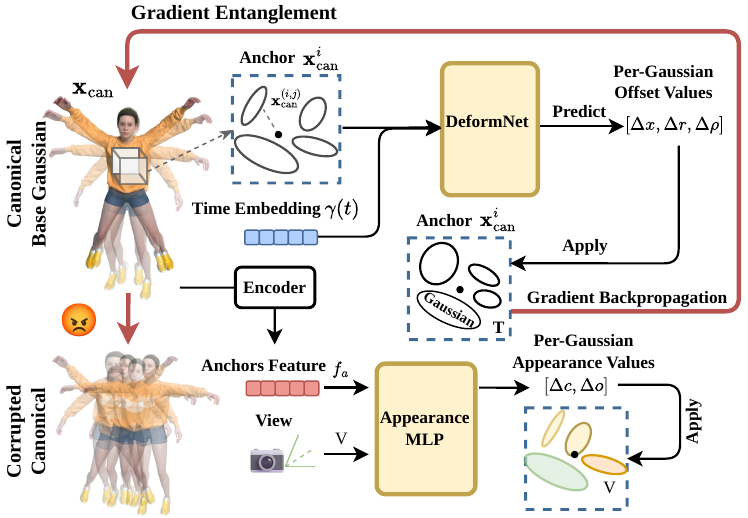}
  \caption{\textbf{Gradient entanglement.} Without architectural isolation, deformation gradients propagate back into canonical geometry through the shared variable $\mathbf{x}_{\text{can}}$, destabilizing the time-invariant reference.}
  \label{fig:grad_entangle}
\end{figure}
\subsection{Dynamic 3D Gaussian Splatting}
Dynamic extensions of 3D-GS, building on earlier non-rigid reconstruction paradigms~\cite{newcombe2015dynamicfusion,innmann2016volumedeform}, introduce time-conditioned
deformation fields mapping each observed frame to a shared
canonical representation~\cite{yang2024deformablethreedga,wu2024fourdgaussiansplat,huang2024scgs,lin2024gaussianflow,luiten2024dynamicthreedgauss,yan2025streetgaussians}. Per-Gaussian approaches operate
without structural priors~\cite{luiten2024dynamicthreedgauss,yang2024deformablethreedga,wu2024fourdgaussiansplat,yang2024gs4d}: Deformable-3DGS~\cite{yang2024deformablethreedga}
learns independent geometric offsets per gaussian conditioned
on position and time, preserving per-gaussian motion degrees
of freedom but suffering from uncontrolled Gaussian growth; 4D-GS~\cite{wu2024fourdgaussiansplat}
encodes dynamic scenes as a unified spatiotemporal Gaussian
field, achieving compact and efficient rendering but coupling
spatial and temporal degrees of freedom within each gaussian.
Anchor-based methods impose structural regularity: SC-GS~\cite{huang2024scgs}
propagates sparse control-node deformations to neighboring
Gaussians via interpolation, imposing structural regularity but
enforcing a single deformation signal per neighborhood, which
suppresses fine-grained local motion.
Gaussian-Flow~\cite{lin2024gaussianflow} parameterizes per-Gaussian motion with polynomial trajectories, achieving temporal smoothness but lacking explicit canonical-deformation decoupling.
4D Scaffold-GS~\cite{cho2026fourdscaffoldgs} extends the Scaffold-GS
anchor hierarchy to dynamic scenes with a dynamic-aware
anchor-growing strategy that densifies anchors in moving regions,
achieving efficient high-fidelity reconstruction, yet it folds
temporal modeling into the shared anchor features rather than
isolating canonical geometry from deformation.
MoDec-GS~\cite{Kwak_2025_CVPR} decomposes scene motion in a
global-to-local manner with temporal interval adjustment to obtain
a compact representation, but its deformation is still optimized
jointly with canonical geometry through a shared gradient pathway.
Despite these advances, a common structural limitation persists:
canonical geometry, the deformation field, and appearance are
optimized through a shared gradient pathway without explicit
module isolation~\cite{yang2024deformablethreedga,wu2024fourdgaussiansplat,huang2024scgs,lin2024gaussianflow,yang2024gs4d}. Deformation gradients continuously perturb
the canonical representation, while temporal photometric
variations introduce appearance gradients that propagate into
canonical geometry through the shared anchor feature~\cite{gao2022dycheck,liu2023rodynrf,liu2025dgmesh}. Both failure
modes share this common source and motivate the design of
GrainGS.

\begin{figure}[t]
  \centering
  \includegraphics[width=\linewidth]{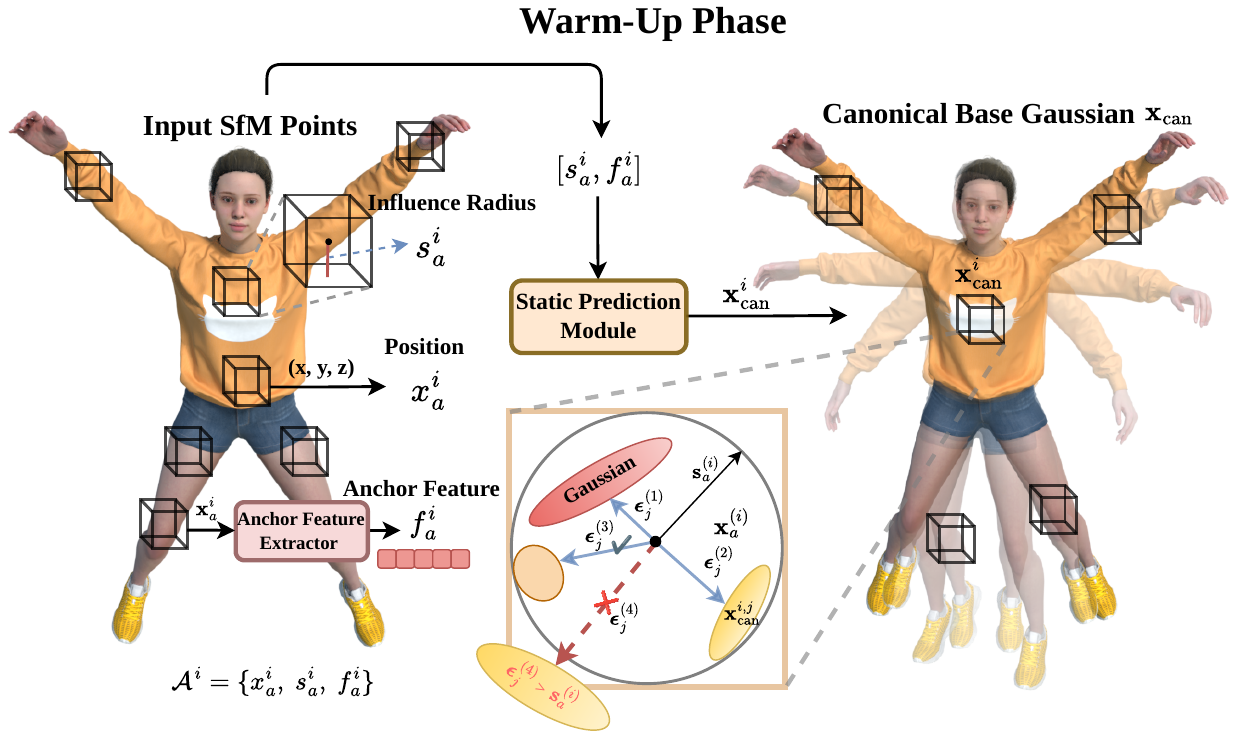}
  \caption{\textbf{Static warm-up.} For the first $T_w$ iterations only the canonical scaffold is optimized, establishing a time-invariant geometric foundation before temporal modules activate.}
  \label{fig:warmup}
\end{figure}

\section{Method}
\label{sec:method}
\begin{figure*}[t!]
  \centering
  \includegraphics[width=0.95\textwidth]{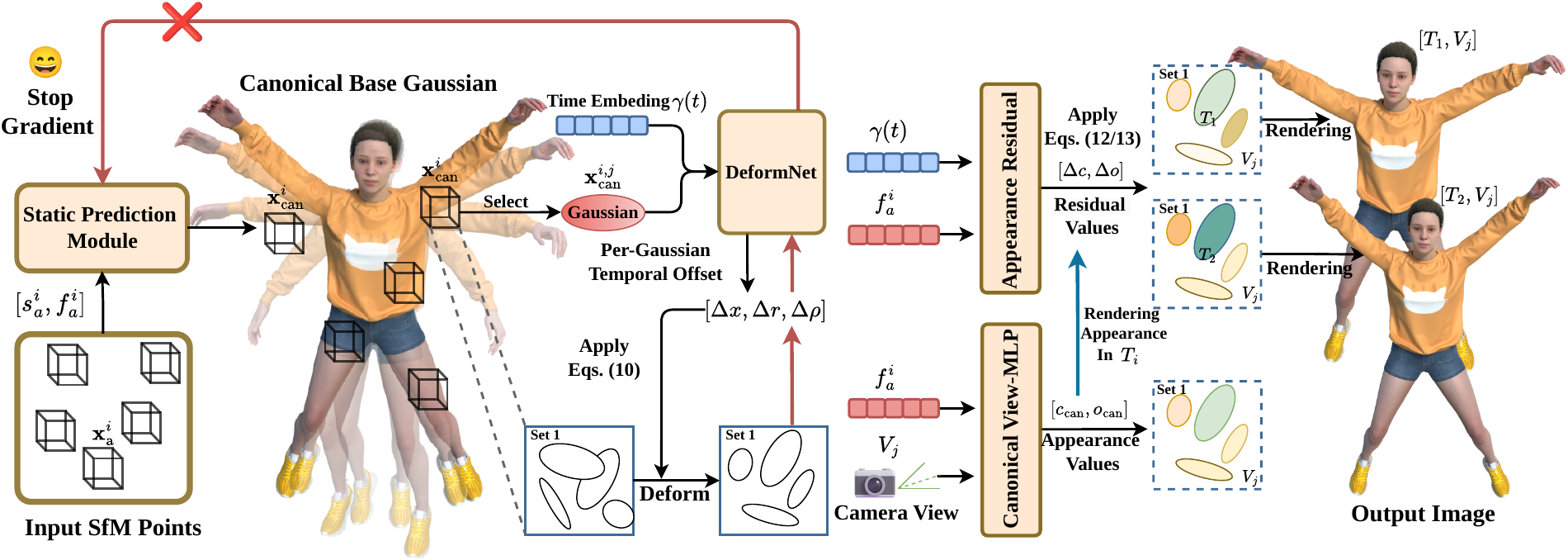}
  \caption{\textbf{Overview of GrainGS.} The pipeline constructs a canonical anchor scaffold, applies per-Gaussian deformation via a time-conditioned DeformNet with stop-gradient isolation, and models appearance through a canonical-residual decomposition.}
  \label{fig:pipeline}
\end{figure*}
\subsection{Overview}
\label{sec:prelim}

3D Gaussian Splatting (3D-GS)~\cite{kerbl2023threedgaussianspla} represents a static
scene as a set of $N$ anisotropic Gaussians.
Each Gaussian $\mathcal{G}_i$ is parameterized by a mean position
$\boldsymbol{\mu}_i\!\in\!\mathbb{R}^3$, a covariance matrix
$\Sigma_i = \mathbf{R}_i\mathbf{S}_i\mathbf{S}_i^\top\mathbf{R}_i^\top$
decomposed into rotation $\mathbf{R}_i\!\in\!\mathrm{SO}(3)$ and
diagonal scale $\mathbf{S}_i\!\in\!\mathbb{R}^{3\times 3}$,
a base opacity $\alpha_i\!\in\![0,1]$, and
view-dependent color $\mathbf{c}_i$.
Each Gaussian contributes to the rendered image through a projected
2D covariance $\tilde{\Sigma}_i = \mathbf{J}\mathbf{W}\Sigma_i\mathbf{W}^\top\mathbf{J}^\top$,
where $\mathbf{W}$ is the view transform and $\mathbf{J}$ the affine
approximation of the projective Jacobian~\cite{kerbl2023threedgaussianspla}.
The per-pixel opacity after projection is
\begin{equation}
    \alpha'_i(\mathbf{p}) = \alpha_i \exp\!\Bigl(-\tfrac{1}{2}
      (\mathbf{p}-\tilde{\boldsymbol{\mu}}_i)^\top
      \tilde{\Sigma}_i^{-1}
      (\mathbf{p}-\tilde{\boldsymbol{\mu}}_i)\Bigr),
    \label{eq:proj_opacity}
\end{equation}
where $\tilde{\boldsymbol{\mu}}_i$ is the projected mean.
Pixel color is obtained by front-to-back $\alpha$-blending after
tile-based depth sorting~\cite{kerbl2023threedgaussianspla}:
\begin{equation}
    \hat{C} = \sum_{i=1}^{N} \mathbf{c}_i\,\alpha'_i(\mathbf{p})
      \prod_{j=1}^{i-1}\!\bigl(1 - \alpha'_j(\mathbf{p})\bigr).
    \label{eq:splat}
\end{equation}
GrainGS extends this static representation to dynamic
scenes through three modules (Fig.~\ref{fig:pipeline}):
a \emph{canonical} anchor scaffold that defines a time-invariant
geometric reference (i.e., a rest-state coordinate frame to which all
temporal deformations are referenced), a per-Gaussian deformation
module with stop-gradient isolation, and a decoupled appearance model.

\subsection{Canonical Anchor-Gaussian Scaffold}
\label{sec:canonical}

Existing dynamic 3D-GS methods suffer from
gradient entanglement partly because they lack a stable geometric
reference: the canonical geometry and the deformation field are
updated through a shared gradient pathway, causing mutual
interference~\cite{yang2024deformablethreedga,wu2024fourdgaussiansplat,huang2024scgs,lin2024gaussianflow}.
To provide such a reference, we adopt a hierarchical anchor
structure that defines a fixed rest-state representation before
any deformation is introduced~\cite{lu2024scaffoldgs}.

\textbf{Anchor initialization.}
An \emph{anchor} is a sparse 3D control point associated with a local support region, from which a set of child Gaussians are generated to
represent the local scene geometry and appearance. Following
Scaffold-GS~\cite{lu2024scaffoldgs}, we initialize anchor positions from
the Structure-from-Motion (SfM) sparse point cloud, followed by voxel
downsampling at resolution $v$ to obtain a spatially uniform anchor
distribution.
Each anchor $i$ is parameterized by a learnable tuple
\begin{equation}
    \mathcal{A}_i = \bigl\{
      \mathbf{x}_{a}^{(i)} \in \mathbb{R}^3,\;\;
      \mathbf{s}_{a}^{(i)} \in \mathbb{R}^3,\;\;
      \mathbf{f}_{a}^{(i)} \in \mathbb{R}^{d_f}
    \bigr\},
\end{equation}
where $\mathbf{x}_{a}^{(i)}$ is the anchor position,
$\mathbf{s}_{a}^{(i)}$ controls the local generation
radius, and $\mathbf{f}_{a}^{(i)}$ is a learnable
per-anchor feature embedding of dimension $d_f$
that encodes local geometric and appearance
information (Fig.~\ref{fig:pipeline}); it is randomly initialized and jointly optimized
with all other parameters during training.

\textbf{Canonical Gaussian generation.}
Each anchor produces $k$ child Gaussians.
The $j$-th child of anchor $i$ carries a learnable offset
$\boldsymbol{\epsilon}_j^{(i)}\!\in\!\mathbb{R}^3$; its
canonical position is
\begin{equation}
    \mathbf{x}_{\text{can}}^{(i,j)}
      = \mathbf{x}_{a}^{(i)}
        + \mathbf{s}_{a}^{(i)} \odot \boldsymbol{\epsilon}_j^{(i)}.
    \label{eq:x_can}
\end{equation}
A shared Static MLP predicts the remaining canonical geometric
attributes for each child Gaussian independently
from the anchor feature and the corresponding
child offset~\cite{lu2024scaffoldgs}:
\begin{equation}
    \bigl(\mathbf{r}_{\text{can}}^{(i,j)},\;
      \boldsymbol{\rho}_{\text{can}}^{(i,j)}\bigr)
      = \mathrm{StaticMLP}\!\left(
          \mathbf{f}_{a}^{(i)},\; \boldsymbol{\epsilon}_j^{(i)}
        \right),
    \label{eq:static_mlp}
\end{equation}
where $\mathbf{r}_{\text{can}}^{(i,j)}\!\in\!\mathbb{R}^4$ is a unit
quaternion and $\boldsymbol{\rho}_{\text{can}}^{(i,j)}\!\in\!\mathbb{R}^3$
is the canonical log-scale vector. The corresponding positive canonical
scale is $\mathbf{s}_{\text{can}}^{(i,j)}=
\exp(\boldsymbol{\rho}_{\text{can}}^{(i,j)})$.
Because different children carry different offsets
$\boldsymbol{\epsilon}_j^{(i)}$, the Static MLP produces distinct
attributes for each of the $k$ Gaussians within the same anchor,
yielding a per-Gaussian canonical representation.
Together, Eqs.~\eqref{eq:x_can}--\eqref{eq:static_mlp} define a
time-independent canonical representation that serves as the geometric
foundation for deformation.

\textbf{Static warm-up.}
To ensure that this canonical scaffold is
established before any deformation gradient can influence it,
we adopt a two-phase training schedule
(Fig.~\ref{fig:warmup}).
For the first $T_w$ iterations, only the anchor parameters
$\{\mathcal{A}_i\}$ and the Static MLP are optimized; the DeformNet
and Appearance Residual Field remains frozen.
Training images are sampled uniformly across all timestamps, so the
canonical scaffold must account for the full spatial extent of the
object's motion range rather than converging to a
single-frame configuration.
While warm-up strategies are common in dynamic
3D-GS pipelines~\cite{yang2024deformablethreedga,wu2024fourdgaussiansplat}, their
typical purpose is to bootstrap point-cloud initialization from
SfM. Our warm-up serves a different goal: establishing a
geometrically stable canonical scaffold before deformation
gradients are introduced, so that the subsequent gradient
decoupling operates on a reliable geometric reference.
When the deformation network activates at iteration $T_w\!+\!1$,
it operates on a geometrically stable scaffold and learns
incremental temporal corrections, rather than being
required to reconstruct geometry and model motion simultaneously.

\textbf{Adaptive anchor growing and pruning.}
A fixed scaffold can under-allocate capacity near fine geometry or persistently hard regions, so we adapt the canonical scaffold
with gradient-driven anchor growing. For each anchor-offset pair $(i,j)$, we
accumulate the screen-space position-gradient norm $G_{i,j}$ and valid
observation count $D_{i,j}$. The average gradient
$\bar{g}_{i,j}=G_{i,j}/(D_{i,j}+\epsilon)$ indicates the expected loss
reduction from moving the corresponding local Gaussian. An offset is selected
as a growing candidate when it is sufficiently observed and its average
gradient exceeds a threshold:
\begin{equation}
    D_{i,j} > \tau_{\mathrm{obs}}, \qquad
    \bar{g}_{i,j} \geq \tau_g .
    \label{eq:anchor_growing_candidate}
\end{equation}
Candidate positions, derived from anchor centers and local offsets, are quantized on a multi-resolution voxel grid. A new anchor is inserted only if the corresponding voxel is unoccupied, thereby preventing redundant growth. Each new anchor inherits the parent features and is initialized with low opacity and zero local offsets before joint optimization with the scaffold. To control model complexity, well-observed anchors with consistently low opacity are pruned. For dynamic scenes, pruning is deferred until the deformation trajectories have stabilized.


\subsection{Gaussian-Level Deformation}
\label{sec:deform}

With the canonical scaffold established, the
remaining challenge is to model per-frame motion without
re-introducing gradient entanglement.
We predict independent deformation offsets for each Gaussian
while using a stop-gradient operator to block gradient flow
back into the canonical scaffold.

\textbf{Per-Gaussian temporal offsets.}
After the warm-up phase, a DeformNet predicts independent per-Gaussian
geometric offsets conditioned on canonical position and time.
The canonical position is detached from the
computational graph via a stop-gradient operator
$\mathrm{sg}(\cdot)$ before being passed to the DeformNet:
\begin{equation}
    \bigl(\Delta\mathbf{x},\;\Delta\mathbf{r},\;
      \Delta\boldsymbol{\rho}\bigr)
      = \mathrm{DeformNet}\!\left(
          \mathrm{sg} \!\bigl(\mathbf{x}_{\text{can}}^{(i,j)}\bigr),\;
          \gamma(t)
        \right),
    \label{eq:deformnet}
\end{equation}
where $\gamma(t)$ is the positional encoding of timestamp $t$~\cite{tancik2020fourier}.
Unlike anchor-based methods that broadcast a single per-anchor
deformation to all child Gaussians, Eq.~\eqref{eq:deformnet} produces
an independent offset for each Gaussian, conditioned
on its own canonical position~\cite{huang2024scgs}.
This per-Gaussian formulation allows the
representation to capture locally varying motion within each anchor
neighborhood, which uniform anchor-broadcast deformation cannot
express.

\textbf{Stop-gradient isolation.}
The operator $\mathrm{sg}(\cdot)$ in Eq.~\eqref{eq:deformnet} blocks
gradient flow from the rendering loss back into the canonical scaffold through the DeformNet.
Without this isolation, deformation gradients
propagate into the Static MLP, causing the canonical geometry to drift
in response to frame-specific motion rather than maintaining a
time-invariant reference.
To see this concretely, consider the deformed position at time $t$, defined as the additive composition
$\mathbf{x}(t)=\mathbf{x}_{\text{can}}+\Delta\mathbf{x}(\mathbf{x}_{\text{can}},\,t)$.
Applying the chain rule to $\mathbf{x}_{\text{can}}$ yields
\begin{equation}
    \frac{\partial \mathcal{L}}{\partial \mathbf{x}_{\text{can}}}
    =
    \frac{\partial \mathcal{L}}{\partial \mathbf{x}(t)}
    \!\cdot\!
    \underbrace{
        \left(
            \mathbf{I}
            +
            \frac{\partial \Delta\mathbf{x}}
                 {\partial \mathbf{x}_{\text{can}}}
        \right)
    }_{\partial\,\mathbf{x}(t)\,/\,\partial\,\mathbf{x}_{\text{can}}}
    =
    \underbrace{
        \frac{\partial \mathcal{L}}{\partial \mathbf{x}(t)}
    }_{\text{reconstruction}}
    +\;
    \underbrace{
        \frac{\partial \mathcal{L}}{\partial \mathbf{x}(t)}
        \!\cdot\!
        \frac{\partial \Delta\mathbf{x}}
             {\partial \mathbf{x}_{\text{can}}}
    }_{\text{deformation}},
    \label{eq:grad_entangle}
\end{equation}
where the first term is the direct photometric reconstruction
gradient and the second term back-propagates through the DeformNet
into the Static MLP via the shared variable
$\mathbf{x}_{\text{can}}$.
The latter term couples the two networks: it shifts
$\mathbf{x}_{\text{can}}$ not only to reduce the photometric
error but also to minimize the deformation
$\Delta\mathbf{x}$ that the DeformNet must predict,
which destabilizes the time-invariant reference.
The stop-gradient operator replaces the DeformNet input with
$\mathrm{sg}(\mathbf{x}_{\text{can}})$, which is treated as a
constant during back-propagation.
This sets
${\partial \Delta\mathbf{x}}/{\partial\,\mathrm{sg}(\mathbf{x}_{\text{can}})}
=\mathbf{0}$
and eliminates the second term in Eq.~\eqref{eq:grad_entangle}:
\begin{equation}
    \frac{\partial \mathcal{L}}{\partial \mathbf{x}_{\text{can}}}
    \bigg|_{\text{w/ sg}}
    =
    \frac{\partial \mathcal{L}}{\partial \mathbf{x}(t)},
    \label{eq:grad_decouple}
\end{equation}
so that the canonical scaffold receives only the photometric
reconstruction signal.
This decouples the optimization of the Static MLP from that of
the DeformNet, ensuring that the canonical geometry remains a
stable, time-invariant reference while the DeformNet independently
learns per-frame deformations.
This decoupling directly realizes the gradient
isolation proposed in Sec.~\ref{sec:intro}: the stop-gradient
operator severs the entangled gradient pathway between canonical
geometry and the deformation network.

\textbf{Deformed attributes.}
The final geometric attributes at time $t$ are composed from canonical
and temporal terms:
\begin{equation}
    \left\{
    \begin{aligned}
      \mathbf{x}(t) &= \mathbf{x}_{\text{can}} + \Delta\mathbf{x}(t), \\
      \mathbf{r}(t) &= \mathbf{r}_{\text{can}} \otimes \Delta\mathbf{r}(t), \\
      \mathbf{s}(t) &= \exp\!\Bigl(\boldsymbol{\rho}_{\text{can}} +
        \operatorname{clip}\!\bigl(\Delta\boldsymbol{\rho}(t),-2,2\bigr)\Bigr).
    \end{aligned}
    \right.
    \label{eq:deformation}
\end{equation}
where $\otimes$ denotes quaternion multiplication. The clipping and
exponential operations are applied element-wise, ensuring positive scales for all
$\Delta\boldsymbol{\rho}\!\in\!\mathbb{R}^3$.

\subsection{Decoupled Appearance Model}
\label{sec:appearance}

The stop-gradient operator in
Sec.~\ref{sec:deform} prevents deformation gradients from
reaching canonical geometry, but it does not block a second
source of interference: temporal photometric variations.
In dynamic sequences, moving shadows, specular
highlights, and exposure changes can be absorbed by the
deformation field as spurious geometric offsets if appearance and
geometry share the same gradient pathway.
To close this remaining pathway, we
decompose appearance into a time-invariant
canonical branch and a regularized temporal residual branch.

\textbf{Canonical View-MLP.}
A pair of time-independent heads predicts canonical color and a neural
opacity score from the anchor feature and the viewing direction $\mathbf{v}$:
\begin{equation}
    \mathbf{c}_{\text{can}} =
      \sigma\!\left(F_c(\mathbf{f}_{a}^{(i)},\mathbf{v})\right),\quad
    o_{\text{can}} =
      \tanh\!\left(F_o(\mathbf{f}_{a}^{(i)},\mathbf{v})\right).
    \label{eq:canonical_app}
\end{equation}
Since $F_c$ and $F_o$ receive no temporal
input, $\mathbf{c}_{\text{can}}$ and $o_{\text{can}}$ can
only represent time-invariant, view-dependent appearance.
Any frame-specific photometric change (e.g., a shadow appearing
at time $t$) cannot be explained by this branch and is therefore
left to the residual branch described below.
In particular, $o_{\text{can}}$ is a neural opacity score rather than
a directly normalized alpha value. It is predicted by this branch
rather than the Static MLP, since the score is
view-dependent and must remain decoupled from geometric
attributes.

\mbox{\textbf{Appearance Residual Field.}}
A lightweight time-conditioned MLP predicts bounded color and opacity-score
residuals from the anchor feature and positional encoding of
time~\cite{tancik2020fourier}:
\begin{equation}
    \bigl(\Delta\mathbf{c}(t),\Delta o(t)\bigr)
      = \eta\tanh\!\left(F_{\mathrm{res}}
        (\mathbf{f}_{a}^{(i)},\gamma(t))\right).
    \label{eq:res_app}
\end{equation}
We set the residual scale to $\eta=0.1$ to keep the temporal corrections
small. The final color and opacity score are composed as
\begin{equation}
    \begin{aligned}
      \mathbf{c}(t) &= \operatorname{clip}\!\left(
        \mathbf{c}_{\text{can}}+\Delta\mathbf{c}(t),0,1\right), \\
      o(t) &= o_{\text{can}}+\Delta o(t).
    \end{aligned}
    \label{eq:c_final}
\end{equation}
Only Gaussians with $o_i(t)>0$ are rendered; their effective per-pixel
alpha follows the Gaussian kernel in Eq.~\eqref{eq:proj_opacity} and is
upper-bounded by $0.99$ in the rasterizer. The residual branch absorbs
frame-specific appearance changes, while $\mathcal{L}_{\text{res}}$ in
Sec.~\ref{sec:training} keeps it subordinate to the canonical appearance.

\begin{table*}[!t]
\centering
\caption{
Quantitative comparison on the synthetic D-NeRF
dataset~\cite{pumarola2021dnerf}. PSNR/SSIM/LPIPS are reported for all
scenes; $\uparrow$/$\downarrow$ denote higher/lower is better. Cell
colors indicate \colorbox{best}{\textbf{best}},
\colorbox{second}{\textbf{second best}}, and
\colorbox{third}{\textbf{third best}} results.
}
\label{tab:dnerf}
\resizebox{\textwidth}{!}{%
\begin{tabular}{lcccccccccccc}
\toprule
\multirow{2}{*}{Method}
& \multicolumn{3}{c}{Hell Warrior}
& \multicolumn{3}{c}{Mutant}
& \multicolumn{3}{c}{Hook}
& \multicolumn{3}{c}{Bouncing Balls} \\
\cmidrule(lr){2-4} \cmidrule(lr){5-7} \cmidrule(lr){8-10} \cmidrule(lr){11-13}
& PSNR$\uparrow$ & SSIM$\uparrow$ & LPIPS$\downarrow$
& PSNR$\uparrow$ & SSIM$\uparrow$ & LPIPS$\downarrow$
& PSNR$\uparrow$ & SSIM$\uparrow$ & LPIPS$\downarrow$
& PSNR$\uparrow$ & SSIM$\uparrow$ & LPIPS$\downarrow$ \\
\midrule
D-NeRF~\cite{pumarola2021dnerf}
  & 24.757 & 0.9537 & 0.0664
  & 30.582 & 0.9609 & 0.0391
  & 28.795 & 0.9544 & 0.0623
  & 36.969 & 0.9860 & 0.0401 \\
Tensor4D~\cite{shao2023tensor4d}
  & 27.880 & 0.9718 & 0.0429
  & 32.186 & 0.9720 & 0.0334
  & 29.123 & 0.9598 & 0.0446
  & 38.667 & 0.9924 & 0.0215 \\
4D-GS~\cite{wu2024fourdgaussiansplat}
  & 28.785 & 0.9736 & 0.0361
  & 37.819 & 0.9885 & 0.0157
  & 33.006 & 0.9764 & 0.0263
  & 41.020 & 0.9945 & 0.0143 \\
D-3DGS~\cite{yang2024deformablethreedga}
  & 32.092 & 0.9827 & 0.0271
  & 38.124 & 0.9893 & 0.0166
  & 34.972 & 0.9846 & 0.0195
  & 42.767 & \cellcolor{best}0.9959 & 0.0110 \\
SC-GS~\cite{huang2024scgs}
  & 32.790 & \cellcolor{second}0.9859 & 0.0186
  & \cellcolor{third}39.061 & \cellcolor{second}0.9942 & \cellcolor{second}0.0068
  & \cellcolor{second}36.806 & \cellcolor{second}0.9903 & \cellcolor{best}0.0099
  & \cellcolor{second}43.081 & \cellcolor{second}0.9959 & \cellcolor{best}0.0094 \\
4D-Scaffold-GS~\cite{cho2026fourdscaffoldgs}
  & \cellcolor{best}33.111 & \cellcolor{third}0.9838 & \cellcolor{third}0.0184
  & \cellcolor{best}39.507 & \cellcolor{third}0.9932 & \cellcolor{third}0.0072
  & 35.801 & 0.9881 & 0.0150
  & 42.697 & 0.9953 & 0.0195 \\
MoDec-GS~\cite{Kwak_2025_CVPR}
  & \cellcolor{third}32.801 & 0.9816 & \cellcolor{second}0.0168
  & 38.965 & 0.9906 & 0.0092
  & \cellcolor{third}36.386 & \cellcolor{third}0.9890 & \cellcolor{third}0.0135
  & \cellcolor{third}42.977 & 0.9949 & \cellcolor{third}0.0107 \\
Ours
  & \cellcolor{second}32.932 & \cellcolor{best}0.9875 & \cellcolor{best}0.0162
  & \cellcolor{second}39.447 & \cellcolor{best}0.9945 & \cellcolor{best}0.0066
  & \cellcolor{best}36.888 & \cellcolor{best}0.9913 & \cellcolor{second}0.0108
  & \cellcolor{best}43.492 & \cellcolor{third}0.9956 & \cellcolor{second}0.0095 \\
\midrule
\multirow{2}{*}{Method}
& \multicolumn{3}{c}{Lego}
& \multicolumn{3}{c}{T-Rex}
& \multicolumn{3}{c}{Stand Up}
& \multicolumn{3}{c}{Jumping Jacks} \\
\cmidrule(lr){2-4} \cmidrule(lr){5-7} \cmidrule(lr){8-10} \cmidrule(lr){11-13}
& PSNR$\uparrow$ & SSIM$\uparrow$ & LPIPS$\downarrow$
& PSNR$\uparrow$ & SSIM$\uparrow$ & LPIPS$\downarrow$
& PSNR$\uparrow$ & SSIM$\uparrow$ & LPIPS$\downarrow$
& PSNR$\uparrow$ & SSIM$\uparrow$ & LPIPS$\downarrow$ \\
\midrule
D-NeRF~\cite{pumarola2021dnerf}
  & 21.711 & 0.8470 & 0.1191
  & 30.301 & 0.9638 & 0.0427
  & 32.244 & 0.9766 & 0.0295
  & 31.802 & 0.9727 & 0.0401 \\
Tensor4D~\cite{shao2023tensor4d}
  & 23.710 & 0.8954 & 0.0972
  & 31.855 & 0.9776 & 0.0294
  & 34.830 & 0.9843 & 0.0211
  & 32.239 & 0.9781 & 0.0306 \\
4D-GS~\cite{wu2024fourdgaussiansplat}
  & 25.078 & 0.9371 & 0.0583
  & 34.388 & 0.9855 & 0.0207
  & 38.272 & 0.9902 & 0.0131
  & 35.337 & 0.9856 & 0.0197 \\
D-3DGS~\cite{yang2024deformablethreedga}
  & 24.834 & 0.9292 & 0.0668
  & 37.014 & 0.9902 & 0.0172
  & \cellcolor{third}39.873 & \cellcolor{third}0.9925 & 0.0117
  & 36.828 & \cellcolor{third}0.9887 & 0.0169 \\
SC-GS~\cite{huang2024scgs}
  & 25.176 & \cellcolor{second}0.9437 & 0.0478
  & \cellcolor{second}38.582 & \cellcolor{second}0.9930 & \cellcolor{second}0.0117
  & \cellcolor{second}40.289 & \cellcolor{second}0.9955 & \cellcolor{second}0.0053
  & \cellcolor{second}38.065 & \cellcolor{second}0.9925 & \cellcolor{second}0.0087 \\
4D-Scaffold-GS~\cite{cho2026fourdscaffoldgs}
  & \cellcolor{third}25.215 & 0.9421 & \cellcolor{third}0.0424
  & \cellcolor{third}38.360 & \cellcolor{third}0.9919 & 0.0135
  & 38.560 & 0.9912 & 0.0102
  & 35.520 & 0.9872 & 0.0180 \\
MoDec-GS~\cite{Kwak_2025_CVPR}
  & \cellcolor{best}25.804 & \cellcolor{best}0.9451 & \cellcolor{best}0.0396
  & 38.219 & 0.9916 & \cellcolor{third}0.0134
  & 39.220 & 0.9906 & \cellcolor{third}0.0102
  & \cellcolor{third}37.247 & 0.9875 & \cellcolor{third}0.0114 \\
Ours
  & \cellcolor{second}25.384 & \cellcolor{third}0.9425 & \cellcolor{second}0.0408
  & \cellcolor{best}38.864 & \cellcolor{best}0.9947 & \cellcolor{best}0.0112
  & \cellcolor{best}40.336 & \cellcolor{best}0.9957 & \cellcolor{best}0.0051
  & \cellcolor{best}38.511 & \cellcolor{best}0.9952 & \cellcolor{best}0.0083 \\
\bottomrule
\end{tabular}%
}

\vspace{4pt}

\includegraphics[width=0.96\textwidth]{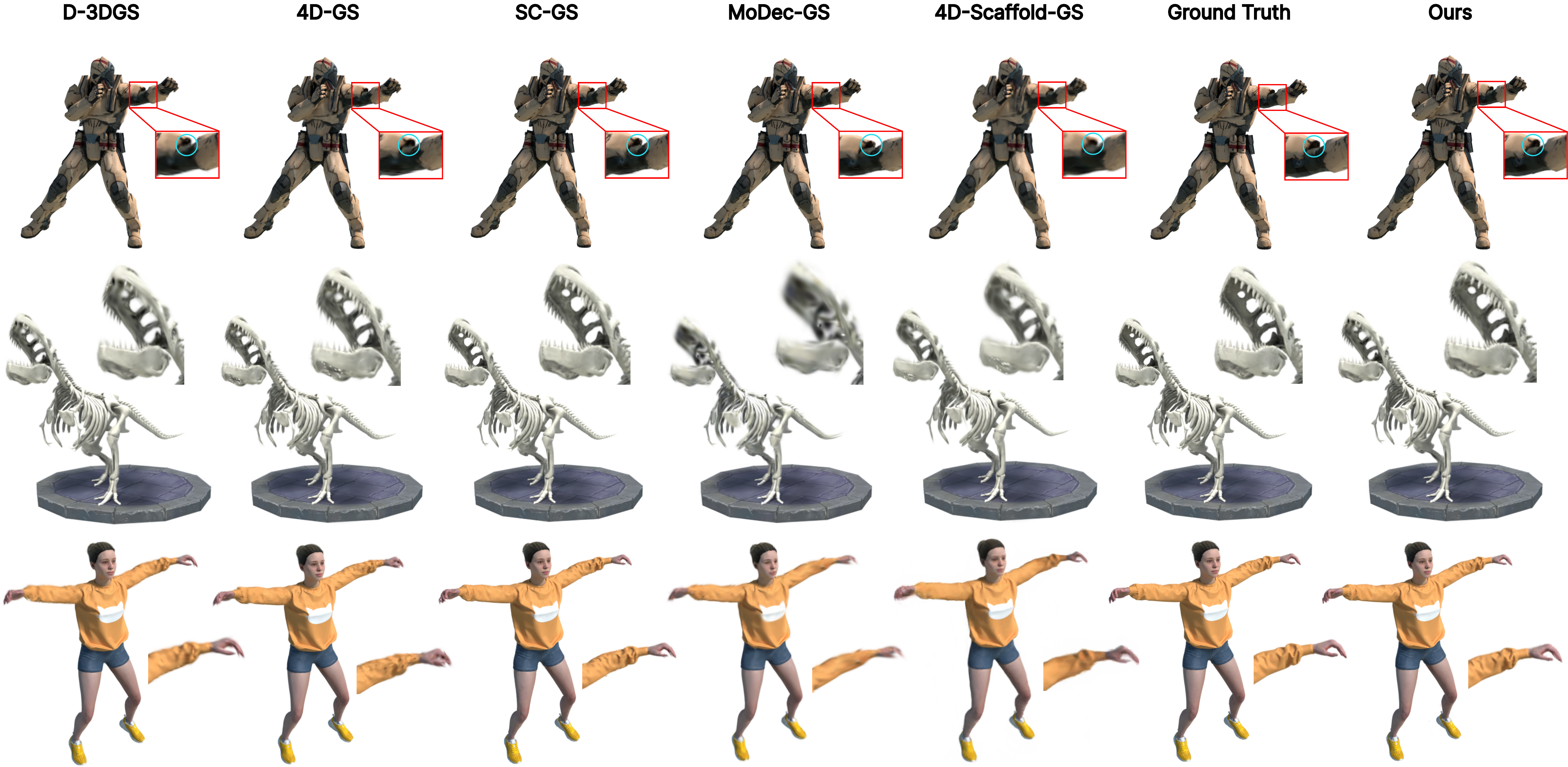}
\captionof{figure}{\textbf{Qualitative comparison on the D-NeRF dataset~\cite{pumarola2021dnerf}.} GrainGS faithfully recovers fine-grained geometric details and sharp boundaries that competing methods blur or distort. Zoom in for best viewing.}
\label{fig:dnerf_visual}
\end{table*}

\begin{table*}[!t]
\centering
\caption{
Quantitative comparison on the real-world DG-Mesh
dataset~\cite{liu2025dgmesh}. PSNR and LPIPS are reported for each
scene; $\uparrow$/$\downarrow$ denote higher/lower is better. Cell
colors indicate \colorbox{best}{\strut best},
\colorbox{second}{\strut second best}, and
\colorbox{third}{\strut third best} results.
}
\label{tab:dgmesh}
\setlength{\tabcolsep}{5pt}
\resizebox{\textwidth}{!}{%
\begin{tabular}{lcccccccccccc}
\toprule
\multirow{2}{*}{Method}
& \multicolumn{2}{c}{Beagle}
& \multicolumn{2}{c}{Bird}
& \multicolumn{2}{c}{Duck}
& \multicolumn{2}{c}{Girlwalk}
& \multicolumn{2}{c}{Horse}
& \multicolumn{2}{c}{Torus2Sphere} \\
\cmidrule(lr){2-3}\cmidrule(lr){4-5}\cmidrule(lr){6-7}
\cmidrule(lr){8-9}\cmidrule(lr){10-11}\cmidrule(lr){12-13}
& PSNR$\uparrow$ & LPIPS$\downarrow$
& PSNR$\uparrow$ & LPIPS$\downarrow$
& PSNR$\uparrow$ & LPIPS$\downarrow$
& PSNR$\uparrow$ & LPIPS$\downarrow$
& PSNR$\uparrow$ & LPIPS$\downarrow$
& PSNR$\uparrow$ & LPIPS$\downarrow$ \\
\midrule
D-NeRF~\cite{pumarola2021dnerf}
  & 36.705 & 0.0396
  & 23.403 & 0.0739
  & 31.857 & 0.0369
  & 30.137 & 0.0247
  & 29.653 & 0.0375
  & 28.803 & 0.0922 \\
Tensor4D~\cite{shao2023tensor4d}
  & 37.073 & 0.0371
  & 25.957 & 0.0502
  & 33.575 & 0.0315
  & 29.951 & 0.0234
  & 30.538 & 0.0342
  & 29.245 & 0.0876 \\
4D-GS~\cite{wu2024fourdgaussiansplat}
  & 42.733 & 0.0208
  & 26.308 & 0.0456
  & 34.889 & 0.0242
  & 33.876 & 0.0151
  & 30.644 & 0.0339
  & 30.056 & 0.0759 \\
D-3DGS~\cite{yang2024deformablethreedga}
  & 42.790 & 0.0212
  & 27.290 & 0.0462
  & 39.026 & 0.0134
  & 35.512 & 0.0128
  & 32.154 & 0.0302
  & 29.823 & 0.0842 \\
SC-GS~\cite{huang2024scgs}
  & \cellcolor{second}44.067 & \cellcolor{best}0.0076
  & \cellcolor{third}28.296 & \cellcolor{best}0.0300
  & \cellcolor{second}39.792 & \cellcolor{best}0.0083
  & \cellcolor{third}37.109 & \cellcolor{best}0.0065
  & \cellcolor{second}33.358 & \cellcolor{second}0.0181
  & \cellcolor{third}30.417 & \cellcolor{second}0.0704 \\
4D-Scaffold-GS~\cite{cho2026fourdscaffoldgs}
  & \cellcolor{third}43.012 & \cellcolor{third}0.0122
  & 27.792 & 0.0463
  & \cellcolor{third}39.076 & \cellcolor{third}0.0098
  & 36.242 & 0.0097
  & 32.762 & \cellcolor{third}0.0194
  & 29.932 & 0.0839 \\
MoDec-GS~\cite{Kwak_2025_CVPR}
  & 42.036 & \cellcolor{third}0.0097
  & \cellcolor{second}28.469 & \cellcolor{third}0.0371
  & 38.765 & 0.0099
  & \cellcolor{second}37.744 & \cellcolor{third}0.0095
  & \cellcolor{third}33.616 & 0.0197
  & \cellcolor{second}30.927 & \cellcolor{third}0.0728 \\
Ours
  & \cellcolor{best}44.433 & \cellcolor{second}0.0087
  & \cellcolor{best}28.577 & \cellcolor{second}0.0325
  & \cellcolor{best}39.800 & \cellcolor{second}0.0097
  & \cellcolor{best}38.013 & \cellcolor{second}0.0091
  & \cellcolor{best}33.794 & \cellcolor{best}0.0172
  & \cellcolor{best}31.613 & \cellcolor{best}0.0657 \\
\bottomrule
\end{tabular}%
}

\vspace{4pt}

\includegraphics[width=0.96\textwidth]{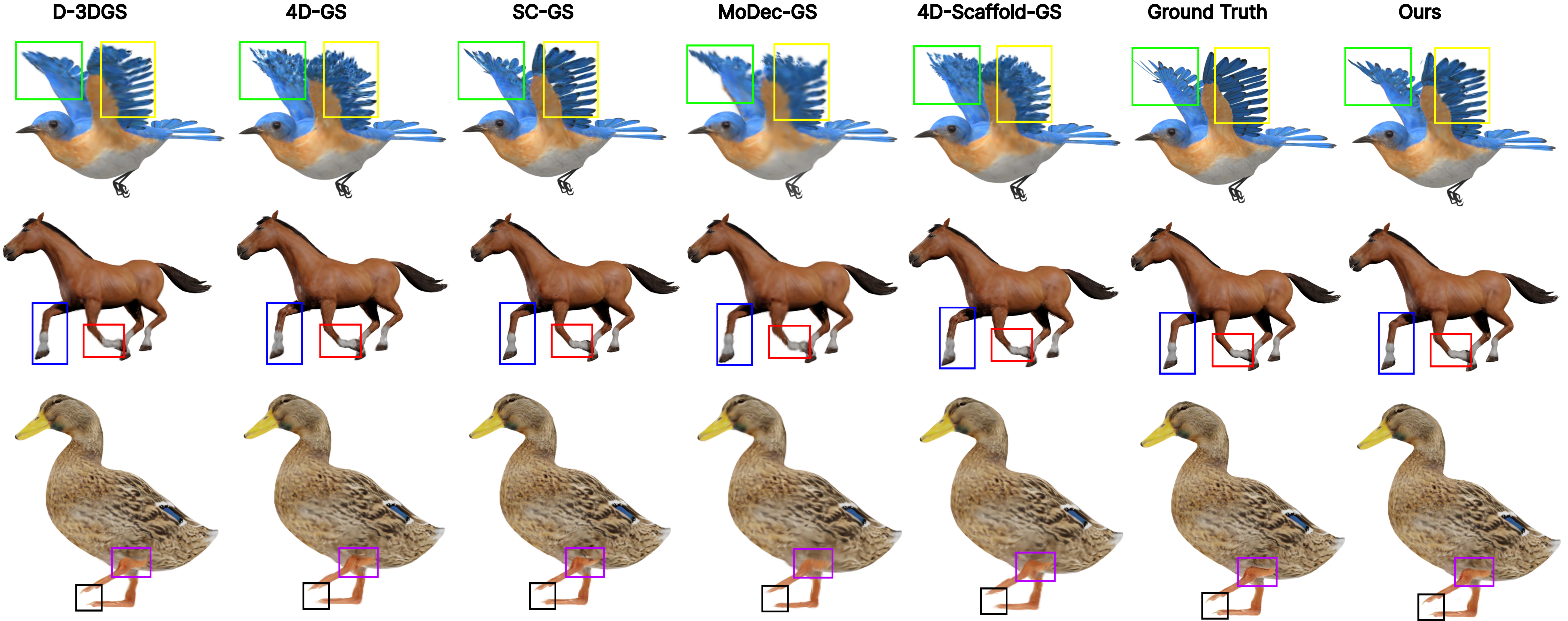}
\captionof{figure}{\textbf{Qualitative comparison on the DG-Mesh dataset~\cite{liu2025dgmesh}.} Across real-world dynamic scenes with non-rigid motion and illumination variation, GrainGS reconstructs cleaner object boundaries, more coherent local geometry, and temporally consistent appearance than competing methods. Zoomed regions highlight improvements in fine structures and texture continuity.}
\label{fig:dgmesh_visual}
\end{table*}

\subsection{Training Objectives and Schedule}
\label{sec:training}

\textbf{Loss function.}
The total training objective is:
\begin{equation}
    \mathcal{L} = \mathcal{L}_{\text{pho}}
      + \lambda_1 \mathcal{L}_{\text{temp}}
      + \lambda_2 \mathcal{L}_{\text{def}}
      + \lambda_3 \mathcal{L}_{\text{res}},
    \label{eq:total_loss}
\end{equation}
where $\mathcal{L}_{\text{pho}}$ is the photometric
reconstruction loss combining $\ell_1$ and D-SSIM terms~\cite{kerbl2023threedgaussianspla}:
\begin{equation}
    \mathcal{L}_{\text{pho}} =
      (1\!-\!\lambda_\text{ssim})\,\|\hat{C} - C\|_1
      + \lambda_\text{ssim}\,\mathcal{L}_{\text{D-SSIM}}.
    \label{eq:pho}
\end{equation}
$\mathcal{L}_{\text{temp}}$ enforces temporal smoothness of
position deformation trajectories by penalizing inter-frame positional offset
discontinuities~\cite{park2021nerfies,li2021nsff,lin2024gaussianflow}:
\begin{equation}
    \mathcal{L}_{\text{temp}} =
      \sum_t \bigl\|\Delta\mathbf{x}(t\!+\!1)
        - \Delta\mathbf{x}(t)\bigr\|_2^2.
    \label{eq:temp}
\end{equation}
$\mathcal{L}_{\text{def}}$ regularizes the magnitude of all
geometric deformation offsets toward zero, biasing the
representation toward the canonical state when photometric
evidence is ambiguous:
\begin{equation}
    \mathcal{L}_{\text{def}} =
      \|\Delta\mathbf{x}\|_2^2
      + \|\Delta\mathbf{r}\|_2^2
      + \|\Delta\boldsymbol{\rho}\|_2^2.
    \label{eq:def}
\end{equation}
$\mathcal{L}_{\text{res}}$ penalizes large appearance residuals to keep the temporal branch subordinate to the canonical base:
\begin{equation}
    \mathcal{L}_{\text{res}} =
      \|\Delta\mathbf{c}\|_2^2
      + \|\Delta o\|_2^2.
    \label{eq:res}
\end{equation}

\textbf{Two-phase training schedule.}
\textbf{(1) Static warm-up} (iterations $1$--$T_w$):
only $\{\mathcal{A}_i\}$ and the Static MLP are optimized
with the loss reduced to $\mathcal{L}_\text{pho}$,
establishing a time-invariant geometric foundation that
spans the full spatial extent of scene motion before
temporal modules are introduced.
\textbf{(2) Joint training} (iterations $T_w\!+\!1$ onward):
all modules are activated; the canonical scaffold receives
gradients exclusively from $\mathcal{L}_\text{pho}$, and
the stop-gradient operator in Eq.~\eqref{eq:deformnet}
ensures that the deformation pathway cannot perturb the
canonical geometry.

The loss terms and training schedule together enforce
independent optimization pathways for canonical geometry,
motion, and appearance, resolving the gradient
entanglement identified in Sec.~\ref{sec:intro}.

\section{Experiments}
\label{sec:experiments}
\subsection{Setup}
\noindent \textbf{\textit{Datasets.}}
We evaluate our method on D-NeRF~\cite{pumarola2021dnerf}, comprising 8 scenes with complex non-rigid motions, and the real-world multi-view DG-Mesh~\cite{liu2025dgmesh}, featuring dynamic captures under natural illumination variation.
\\
\textbf{\textit{Evaluation Metrics.}}
We report PSNR, SSIM, LPIPS, rendering FPS, and total Gaussian count. All experiments run on a single NVIDIA RTX 4090.
\\
\textbf{\textit{Implementation Details.}}
Total training: 30,000 iterations. Warm-up: $T_w = 3{,}000$. Child Gaussians per anchor: $k=10$. Time encoding: sinusoidal positional encoding~\cite{tancik2020fourier} with $L=6$ frequencies. Loss weights: $\lambda_1=0.01$, $\lambda_2=0.001$, $\lambda_3=0.01$, $\lambda_\text{ssim}=0.2$. Anchor-growing statistics start after 500 iterations, and anchor updates are performed every 100 iterations from iteration 1,500 to 20,000. Opacity-based pruning is disabled for the first $80\%$ of the joint training phase to prevent premature removal of Gaussians whose deformation trajectories have not yet been estimated by the DeformNet. 
\subsection{Comparison with State-of-the-Art}
\label{sec:comparison}

We compare GrainGS against seven representative dynamic novel-view-synthesis methods spanning NeRF-based (D-NeRF~\cite{pumarola2021dnerf}, Tensor4D~\cite{shao2023tensor4d}), per-primitive Gaussian (4D-GS~\cite{wu2024fourdgaussiansplat}, Deformable-3DGS (D-3DGS)~\cite{yang2024deformablethreedga}), compact motion-decomposition (MoDec-GS~\cite{Kwak_2025_CVPR}), and anchor-based Gaussian (SC-GS~\cite{huang2024scgs}, 4D-Scaffold-GS~\cite{cho2026fourdscaffoldgs}) paradigms.

\subsubsection{Quantitative Results}

\textbf{D-NeRF dataset~\cite{pumarola2021dnerf}.}
Table~\ref{tab:dnerf} reports per-scene PSNR / SSIM / LPIPS.
GrainGS ranks in the top three for all 24 metric entries across 8 scenes and ranks first or second in 22 entries, achieving a mean PSNR of 36.98\,dB.
It surpasses the strongest average baseline SC-GS (36.73\,dB) by 0.25\,dB and MoDec-GS (36.45\,dB) by 0.53\,dB.
The advantage is most pronounced on scenes with complex localized dynamics:
on \textit{Jumping Jacks}, GrainGS achieves 38.51\,dB versus 38.07\,dB for SC-GS ($+$0.45\,dB) and 36.83\,dB for D-3DGS ($+$1.68\,dB);
on \textit{Mutant}, GrainGS attains the best SSIM (0.9945) and LPIPS (0.0066) and is on par with 4D-Scaffold-GS in PSNR (39.45 vs.\ 39.51\,dB).
These scenes involve rapid limb articulation that demands per-Gaussian deformation granularity;
the uniform deformation bias inherent in anchor-broadcast methods limits SC-GS, while the lack of structural priors in D-3DGS leads to geometric instability.
On scenes with more regular object dynamics, MoDec-GS is particularly competitive on \textit{Lego}; nevertheless, GrainGS remains among the top two on the average D-NeRF metrics and achieves the best LPIPS on the challenging articulated scenes \textit{Mutant}, \textit{T-Rex}, \textit{Stand Up}, and \textit{Jumping Jacks}.

\textbf{DG-Mesh dataset~\cite{liu2025dgmesh}.}
Table~\ref{tab:dgmesh} summarizes PSNR and LPIPS on the 6 real-world DG-Mesh scenes.
GrainGS ranks first in PSNR across all 6 scenes, outperforming the strongest MoDec-GS baseline by 0.27\,dB on \textit{Girlwalk} (38.01 vs.\ 37.74\,dB) and by 0.69\,dB on \textit{Torus2Sphere} (31.61 vs.\ 30.93\,dB).
On \textit{Girlwalk} and \textit{Horse}, where illumination variation is most pronounced, GrainGS outperforms all baselines in PSNR while maintaining competitive LPIPS; it achieves the best LPIPS on \textit{Horse} and \textit{Torus2Sphere}.
This confirms the effectiveness of the canonical-residual appearance decomposition: methods without temporal appearance modeling (\emph{e.g.}, D-3DGS~\cite{yang2024deformablethreedga}, 4D-GS~\cite{wu2024fourdgaussiansplat}) absorb lighting shifts as spurious geometric offsets, degrading perceptual quality.

\textbf{Rendering efficiency.}
Table~\ref{tab:efficiency} reports average FPS and model storage across all D-NeRF scenes.
GrainGS achieves 435.6\,FPS, which is $1.4\times$ faster than D-3DGS (318.2\,FPS) and $1.7\times$ faster than SC-GS (254.2\,FPS).
Against two recent compact dynamic methods, GrainGS renders $1.15\times$ faster than 4D-Scaffold-GS (379.4\,FPS) and $1.29\times$ faster than MoDec-GS (337.6\,FPS).
The anchor-based scaffold constrains Gaussian placement to structurally meaningful locations, yielding a compact representation with a storage of only 4.67\,MB, compared to 32.49\,MB for SC-GS ($7.0\times$ reduction) and 23.68\,MB for 4D-GS ($5.1\times$ reduction).
It also remains $2.8\times$ and $3.5\times$ more compact than 4D-Scaffold-GS (13.30\,MB) and MoDec-GS (16.40\,MB), respectively.
Notably, GrainGS achieves storage competitive with D-3DGS (6.62\,MB) while delivering substantially higher PSNR.

\begin{table}[t]
\centering
\caption{Average rendering speed (FPS) and model storage (MB) on the D-NeRF
dataset~\cite{pumarola2021dnerf}, averaged over all eight scenes;
$\uparrow$/$\downarrow$ indicate that higher/lower is better.
The color of each cell means:
\colorbox{best}{\strut best},
\colorbox{second}{\strut second best},
\colorbox{third}{\strut third best}.}
\label{tab:efficiency}
\small
\setlength{\tabcolsep}{6pt}
\renewcommand{\arraystretch}{1.15}
\begin{tabular*}{\columnwidth}{@{\extracolsep{\fill}}l cc}
\toprule
\textbf{Method} & FPS$\uparrow$ & Storage (MB)$\downarrow$ \\
\midrule
4D-GS~\cite{wu2024fourdgaussiansplat}        & 237.1 & 23.68 \\
D-3DGS~\cite{yang2024deformablethreedga}     & 318.2 & \cellcolor{second}6.62 \\
SC-GS~\cite{huang2024scgs}                   & 254.2 & 32.49 \\
MoDec-GS~\cite{Kwak_2025_CVPR}               & \cellcolor{third}337.6 & 16.40 \\
4D-Scaffold-GS~\cite{cho2026fourdscaffoldgs} & \cellcolor{second}379.4 & \cellcolor{third}13.30 \\
\midrule
Ours                                         & \cellcolor{best}\textbf{435.6} & \cellcolor{best}\textbf{4.67} \\
\bottomrule
\end{tabular*}
\end{table}

\begin{figure}[!b]
  \centering
  \begin{subfigure}{0.48\linewidth}
    \centering
    \includegraphics[width=\linewidth]{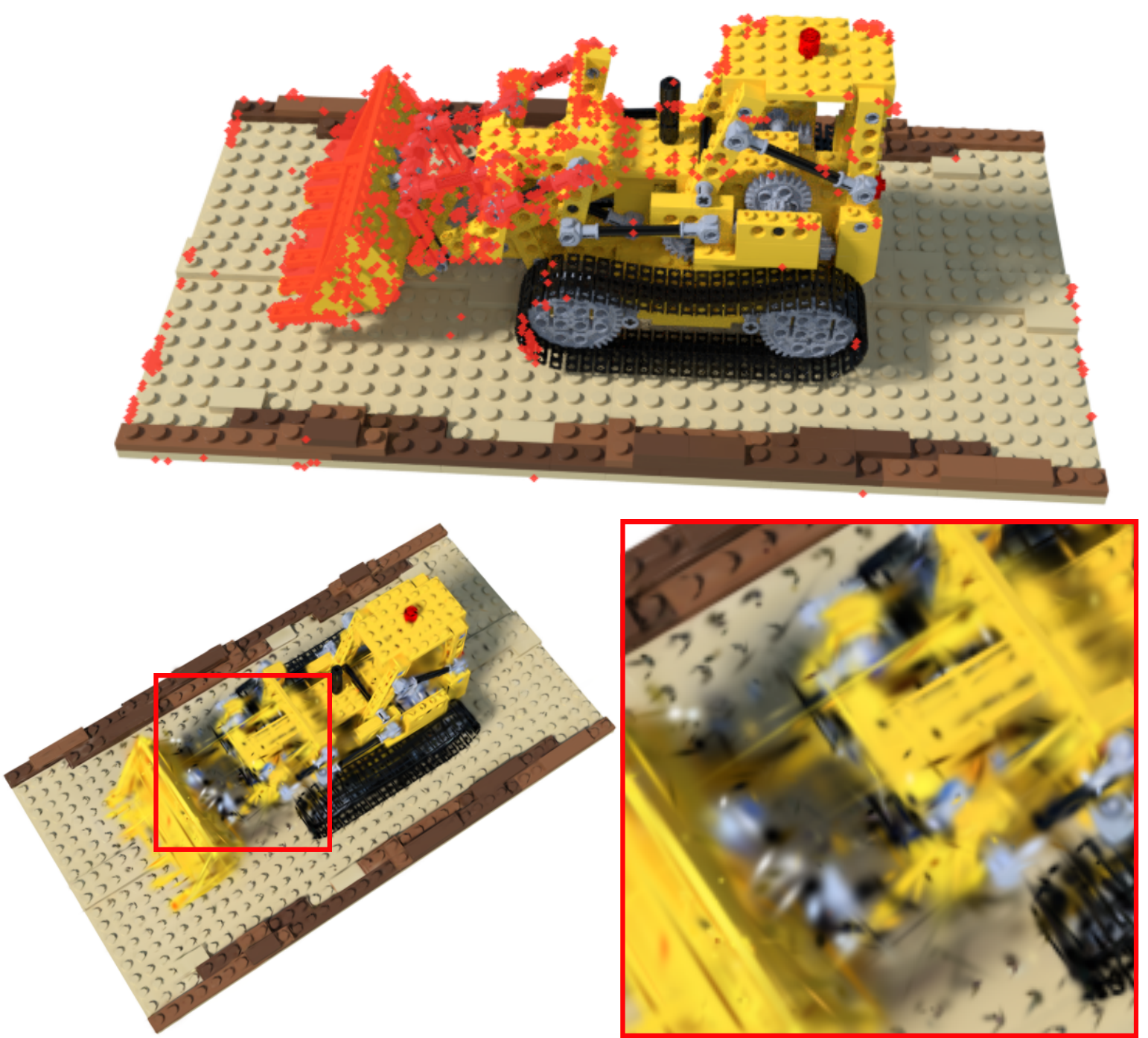}
    \caption{5k iterations}
  \end{subfigure}
  \hfill
  \begin{subfigure}{0.48\linewidth}
    \centering
    \includegraphics[width=\linewidth]{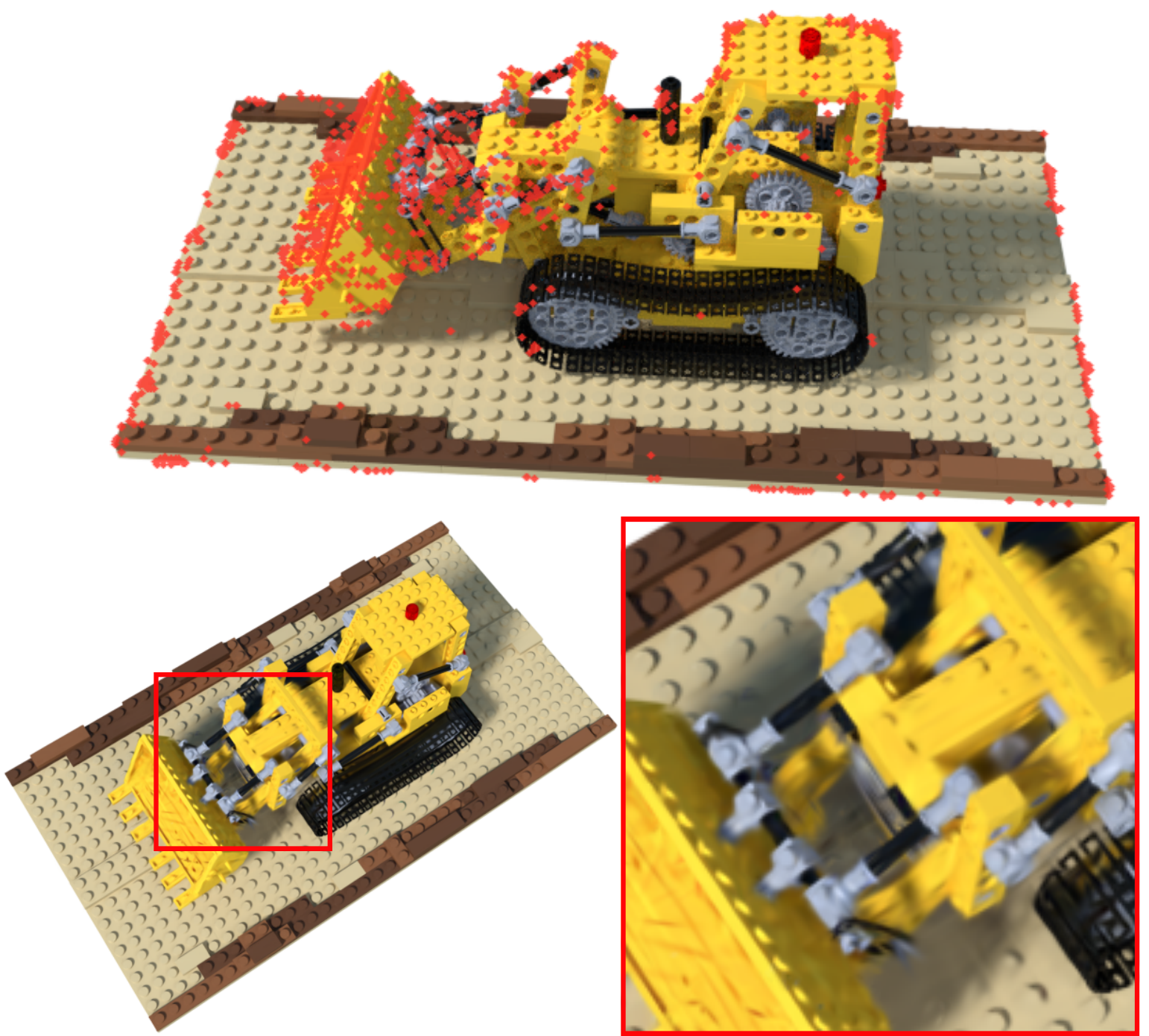}
    \caption{10k iterations}
  \end{subfigure}

  \vspace{2pt}

  \begin{subfigure}{0.48\linewidth}
    \centering
    \includegraphics[width=\linewidth]{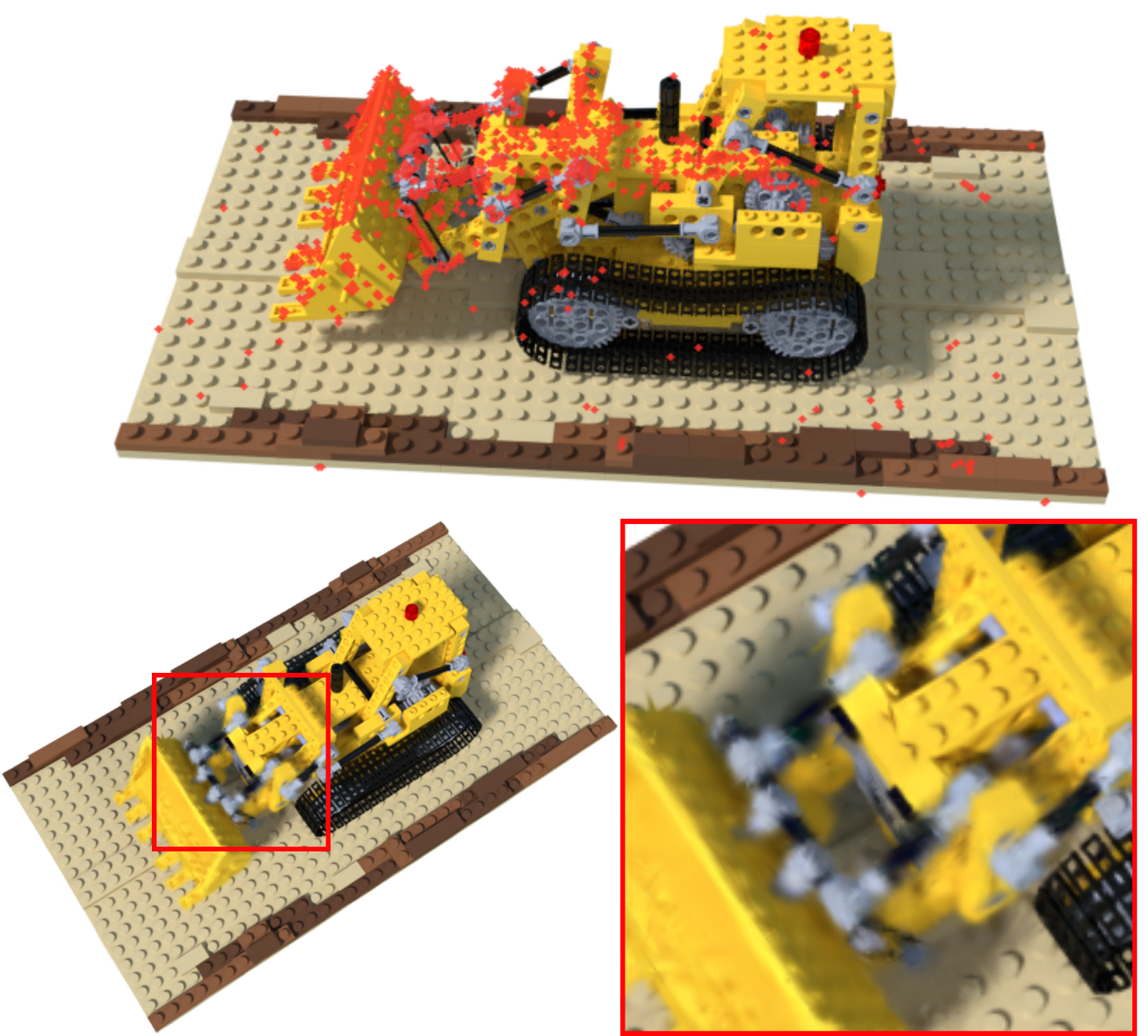}
    \caption{15k iterations}
  \end{subfigure}
  \hfill
  \begin{subfigure}{0.48\linewidth}
    \centering
    \includegraphics[width=\linewidth]{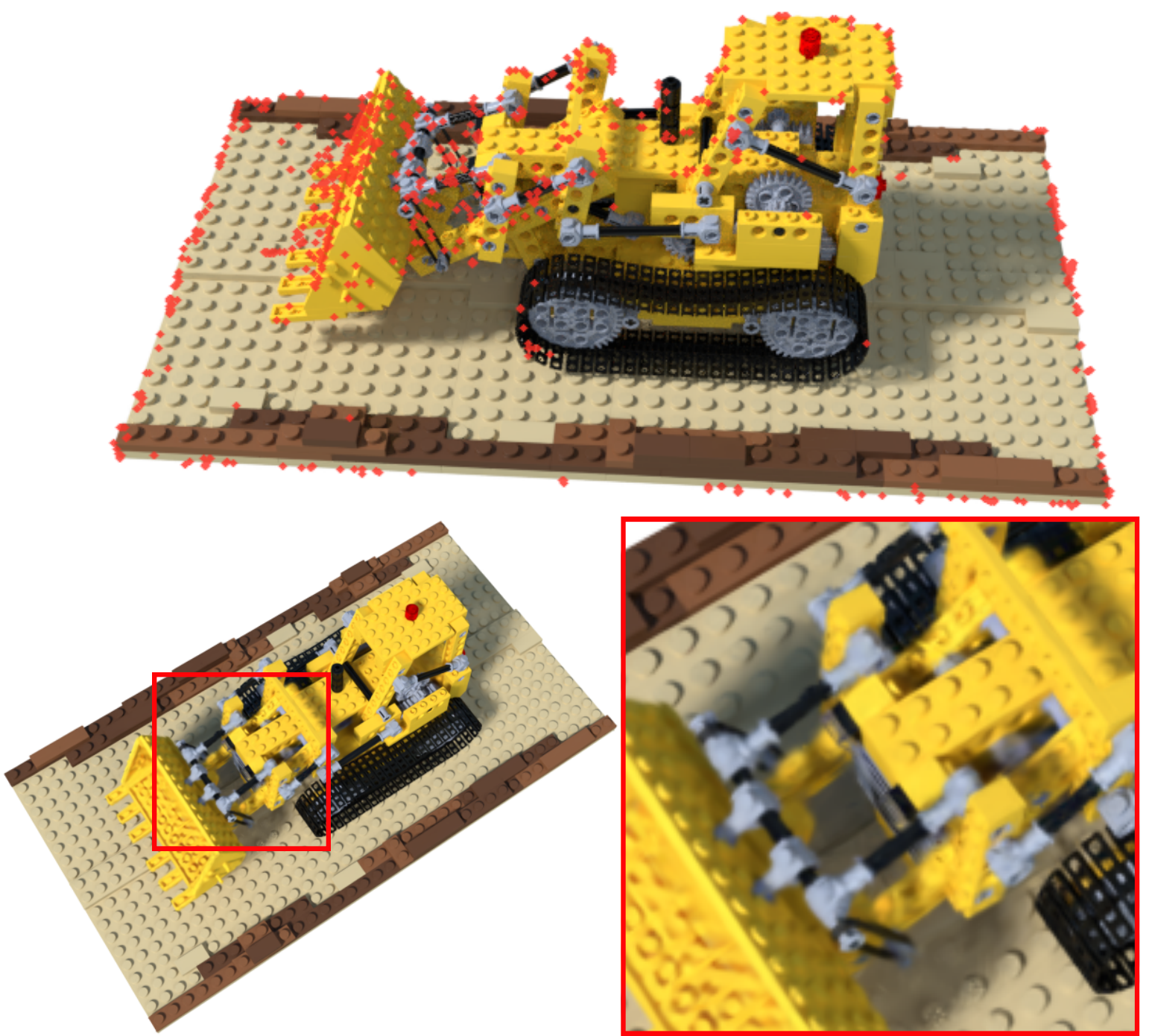}
    \caption{20k iterations}
  \end{subfigure}

  \caption{
  \textbf{Visualization of adaptive anchor growing.}
  Red points denote grown anchors at different training iterations on the D-NeRF \textit{Lego} scene. The scaffold progressively concentrates around geometrically informative regions, increasing local capacity where reconstruction remains difficult while preserving a compact anchor distribution.}
  \label{fig:anchor_growing}
\end{figure}

\textbf{Anchor growing visualization.}
Fig.~\ref{fig:anchor_growing} visualizes the adaptive anchor-growing process on the D-NeRF \textit{Lego} scene. Early in training, anchors are concentrated around high-error regions with incomplete local support. As optimization proceeds, newly inserted anchors progressively cover fine object structures and boundary regions, improving local reconstruction quality while avoiding uniform, scene-wide densification. The scaffold stabilizes by 20k iterations, consistent with the compact storage and high rendering speed reported in Table~\ref{tab:efficiency}.

\subsubsection{Qualitative Results}

Fig.~\ref{fig:dnerf_visual} presents visual comparisons on three representative D-NeRF scenes.
On \textit{Hook} (row~1), the magnified region highlights the arm armor: GrainGS recovers fine-grained structural detail that closely matches the ground truth, while baselines produce over-smoothed or incomplete geometry.
On \textit{Jumping Jacks} (row~2), the inset focuses on finger regions, where GrainGS maintains clear contour structure without artifacts, whereas competing methods exhibit blurred or fragmented boundaries.
On \textit{T-Rex} (row~3), GrainGS reconstructs the dinosaur skeleton without generating spurious geometric structures, and preserves sharp tooth detail at the edges that other methods render as blurred.

Fig.~\ref{fig:dgmesh_visual} shows comparisons on the DG-Mesh dataset~\cite{liu2025dgmesh}.
On \textit{Bird} (row~1), the magnified wing region shows that GrainGS produces sharper feather edges and more accurate color reproduction compared to baselines, which exhibit blurred boundaries and color deviation.
On \textit{Horse} (row~2), GrainGS renders more accurate shadow transitions and correct color continuity at the leg junctions, whereas other methods introduce shadow artifacts or unnatural color discontinuities.
On \textit{Duck} (row~3), GrainGS correctly reconstructs the foot-body junction; all baselines produce structural errors at this connection.
These improvements stem from the canonical-residual appearance decomposition, which prevents temporal illumination variation from corrupting the geometric representation.

\subsection{Ablation Study}
\label{sec:ablation}
We ablate each key component on the D-NeRF dataset~\cite{pumarola2021dnerf} with all other hyperparameters held constant.

\begin{table}[!hbp]
\centering
\caption{Ablation of the stop-gradient mechanism on D-NeRF under otherwise identical settings.}
\label{tab:ablation_stopgrad}
\small
\setlength{\tabcolsep}{6pt}
\setlength{\aboverulesep}{3pt}
\setlength{\belowrulesep}{3pt}
\begin{tabular*}{0.96\columnwidth}{@{\extracolsep{\fill}}l ccc}
\toprule
\textbf{Method} & PSNR$\uparrow$ & SSIM$\uparrow$ & LPIPS$\downarrow$ \\
\midrule
Full Model   & \textbf{36.982} & \textbf{0.9871} & \textbf{0.0136} \\
w/o stop-grad & 35.643 & 0.9812 & 0.0253 \\
\bottomrule
\end{tabular*}
\end{table}

\begin{figure}[h]
  \centering
  \begin{subfigure}{0.35\linewidth}
    \centering
    \includegraphics[width=0.6\linewidth]{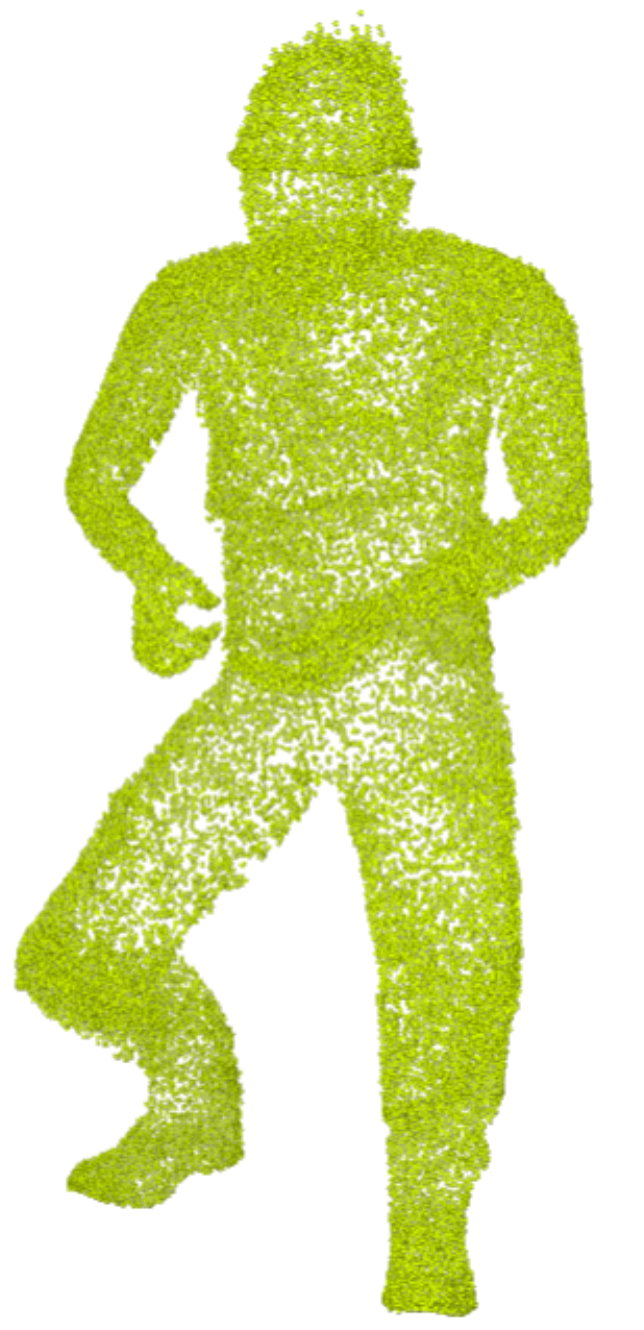}
    \caption{w/ stop-gradient}
  \end{subfigure}
  \hspace{0.04\linewidth}
  \begin{subfigure}{0.35\linewidth}
    \centering
    \includegraphics[width=0.6\linewidth]{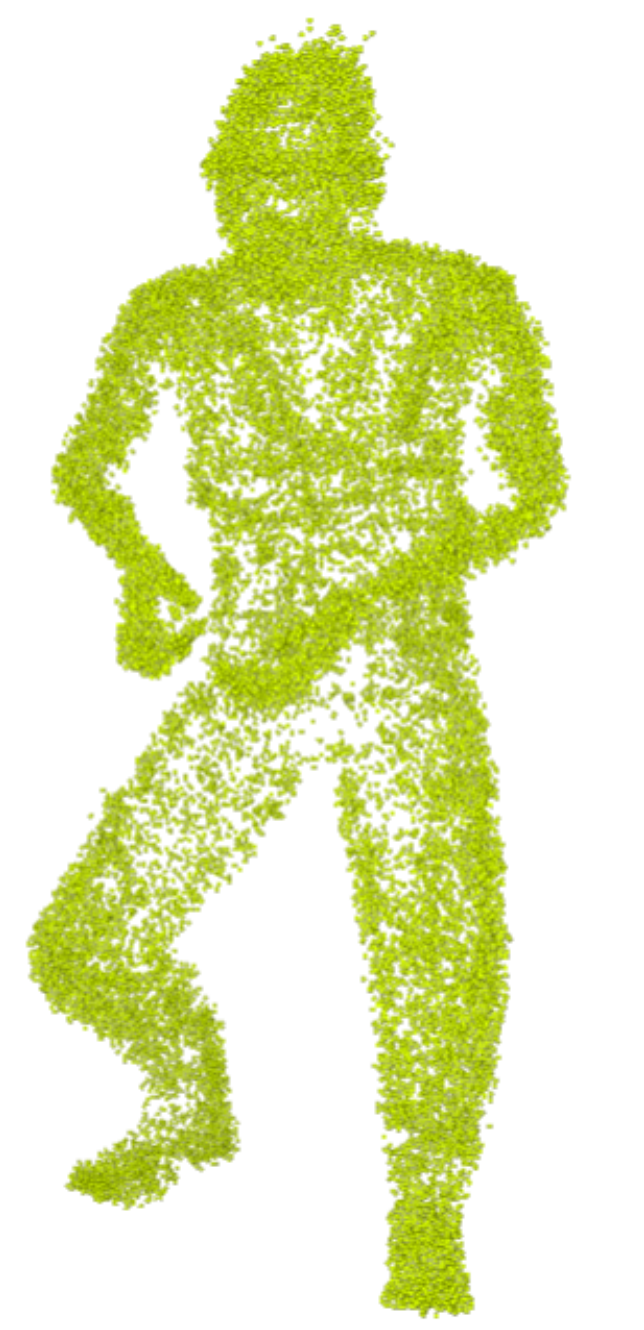}
    \caption{w/o stop-gradient}
  \end{subfigure}
  \caption{
  \textbf{Visual effect of stop-gradient isolation.}
  With stop-gradient, the canonical scaffold produces uniform Gaussians; without it, deformation gradients corrupt the distribution.}
  \label{fig:stopgrad_visual}
\end{figure}

\textbf{Stop-gradient isolation} produces the largest single improvement ($+$1.34\,dB, Table~\ref{tab:ablation_stopgrad}).
Without the stop-gradient operator, deformation gradients propagate into the canonical scaffold, causing the Static MLP to absorb frame-specific motion as permanent geometric distortion.
Fig.~\ref{fig:stopgrad_visual} confirms this effect: with the stop-gradient operator (a), the canonical scaffold produces spatially uniform Gaussian distributions that faithfully cover the object geometry; without it (b), gradient entanglement disrupts the static structure, {yielding} irregular point distributions concentrated in high-motion regions.
This result confirms the gradient-entanglement hypothesis ({Sec.~\ref{sec:intro}}) and demonstrates that explicit gradient isolation is necessary for maintaining a stable canonical reference.

\begin{figure}[!hbp]
  \centering
  \begin{subfigure}{0.43\linewidth}
    \centering
    \includegraphics[width=\linewidth]{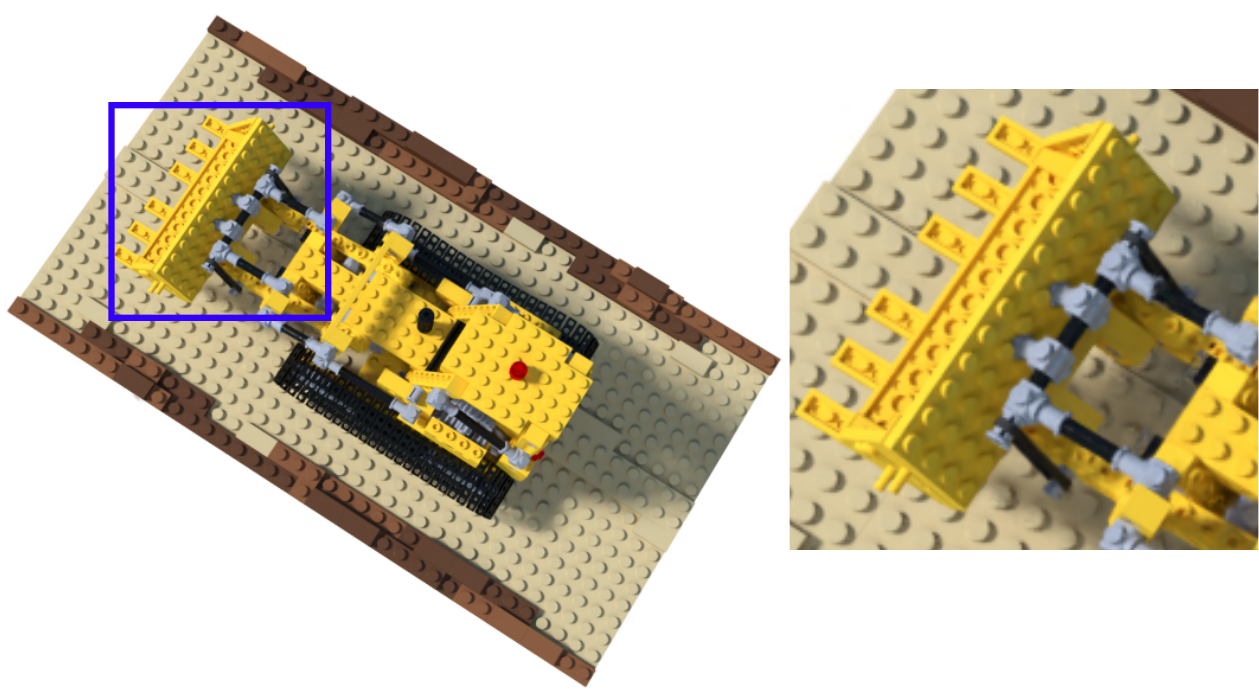}
    \caption{Full Model}
  \end{subfigure}
  \begin{subfigure}{0.43\linewidth}
    \centering
    \includegraphics[width=\linewidth]{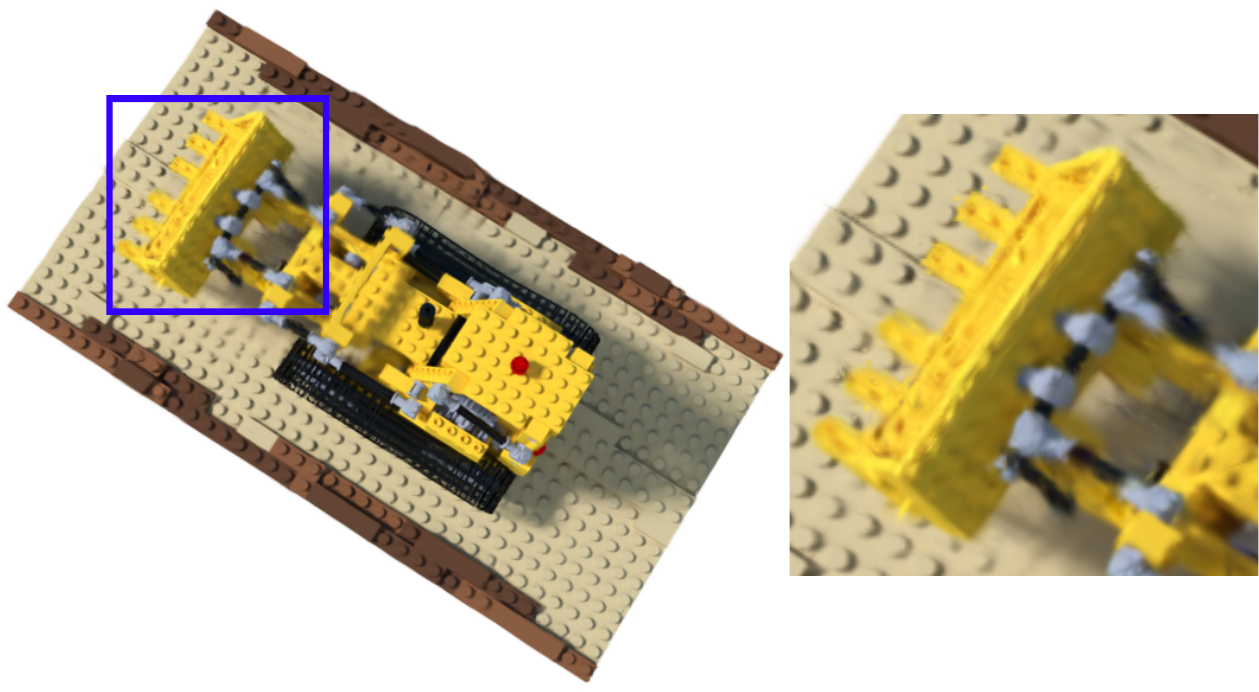}
    \caption{w/o $\mathcal{L}_{\text{temp}}$}
  \end{subfigure}

  \vspace{2pt}

  \begin{subfigure}{0.43\linewidth}
    \centering
    \includegraphics[width=\linewidth]{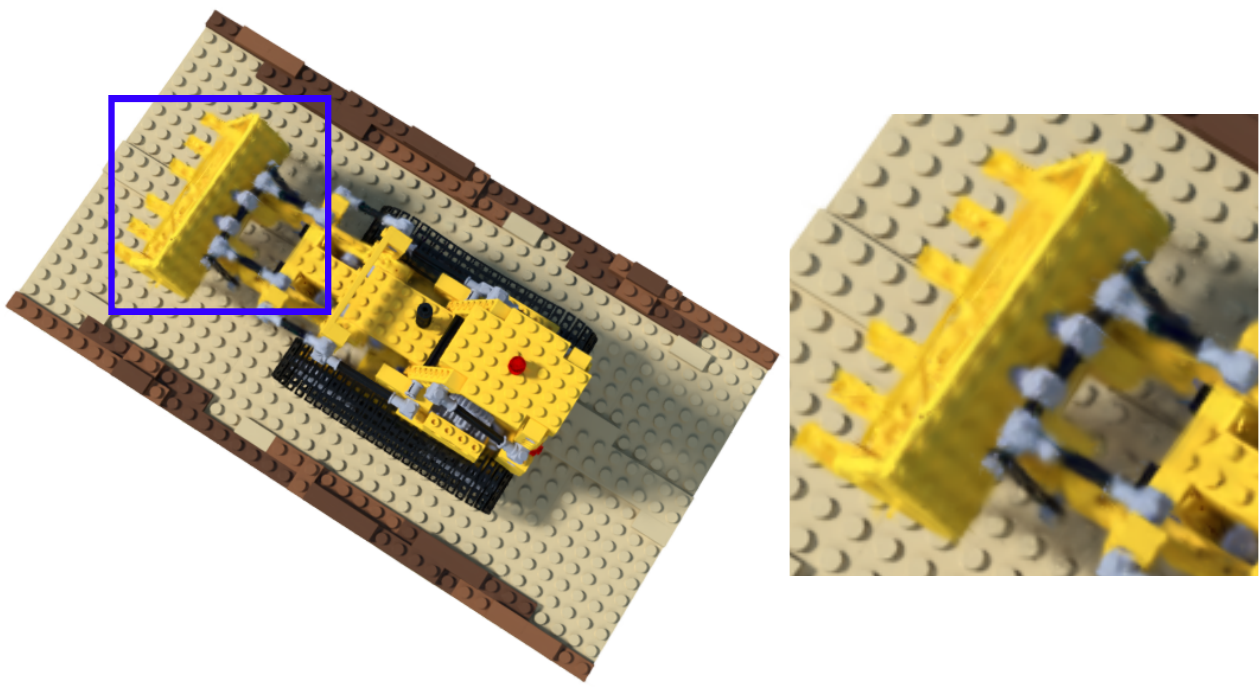}
    \caption{w/o $\mathcal{L}_{\text{def}}$}
  \end{subfigure}
  \begin{subfigure}{0.43\linewidth}
    \centering
    \includegraphics[width=\linewidth]{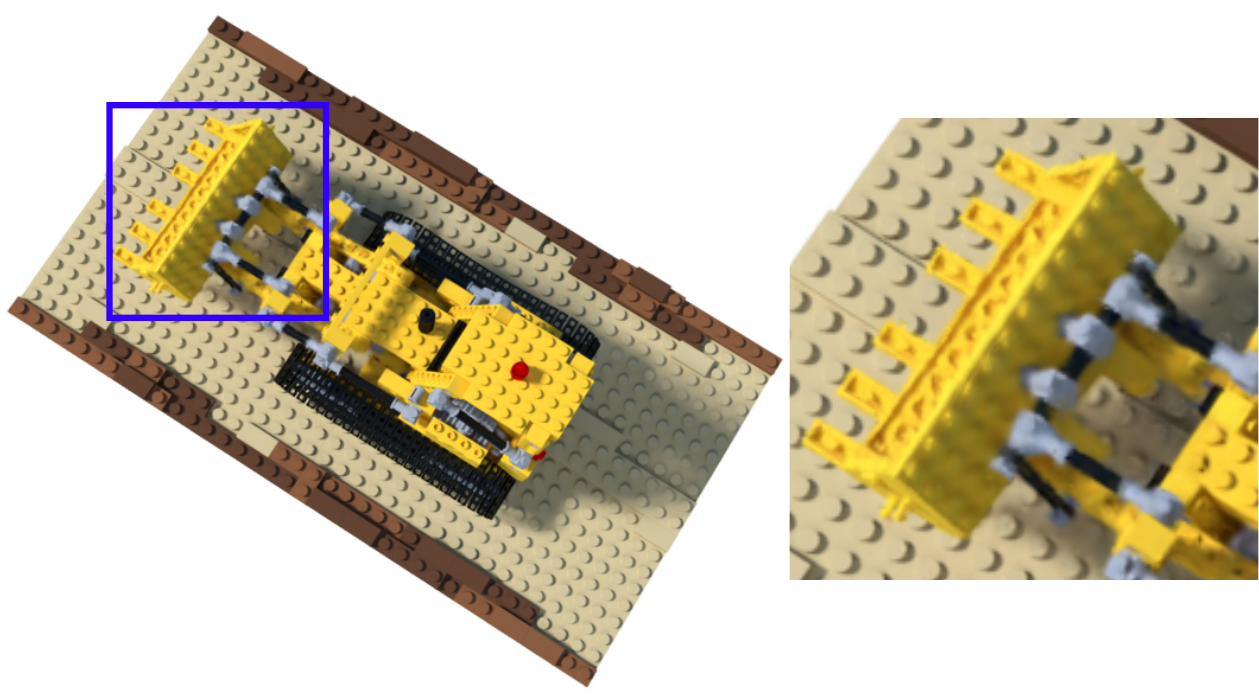}
    \caption{w/o $\mathcal{L}_{\text{res}}$}
  \end{subfigure}
  \caption{
  \textbf{Ablation: loss components.}
  Removing regularization terms leads to progressively degraded quality.}
  \label{fig:loss_ablation}
\end{figure}

\textbf{Loss components.}
Table~\ref{tab:ablation_loss} ablates the three regularization terms, and Fig.~\ref{fig:loss_ablation} shows the corresponding visual results.
The Full Model~(a) produces the sharpest rendering.
Removing $\mathcal{L}_{\text{temp}}$ ((b), $-$0.77\,dB) allows temporally discontinuous deformation trajectories, producing significant blurring across the rendered image.
Removing $\mathcal{L}_{\text{def}}$ ((c), $-$0.52\,dB) permits unconstrained offset magnitudes; the result exhibits noticeable blurring in fine structural details, though less severe than~(b).
Removing $\mathcal{L}_{\text{res}}$ ((d), $-$0.34\,dB) allows the appearance residual branch to override the canonical base, introducing slight blurring in the output.

\begin{table}[!hbp]
\centering
\caption{Ablation of loss components on D-NeRF, showing effects on reconstruction quality under identical settings}
\label{tab:ablation_loss}
\small
\setlength{\tabcolsep}{6pt}
\setlength{\aboverulesep}{3pt}
\setlength{\belowrulesep}{3pt}
\begin{tabular*}{0.96\columnwidth}{@{\extracolsep{\fill}}l ccc}
\toprule
\textbf{Method} & PSNR$\uparrow$ & SSIM$\uparrow$ & LPIPS$\downarrow$ \\
\midrule
Full Model                       & \textbf{36.982} & \textbf{0.9871} & \textbf{0.0136} \\
w/o $\mathcal{L}_{\text{temp}}$  & 36.214 & 0.9829 & 0.0167 \\
w/o $\mathcal{L}_{\text{def}}$   & 36.461 & 0.9856 & 0.0154 \\
w/o $\mathcal{L}_{\text{res}}$   & 36.647 & 0.9861 & 0.0148 \\
\bottomrule
\end{tabular*}
\end{table}

\begin{figure}[!t]
  \centering

  \begin{subfigure}{0.42\linewidth}
    \centering
    \includegraphics[width=\linewidth]{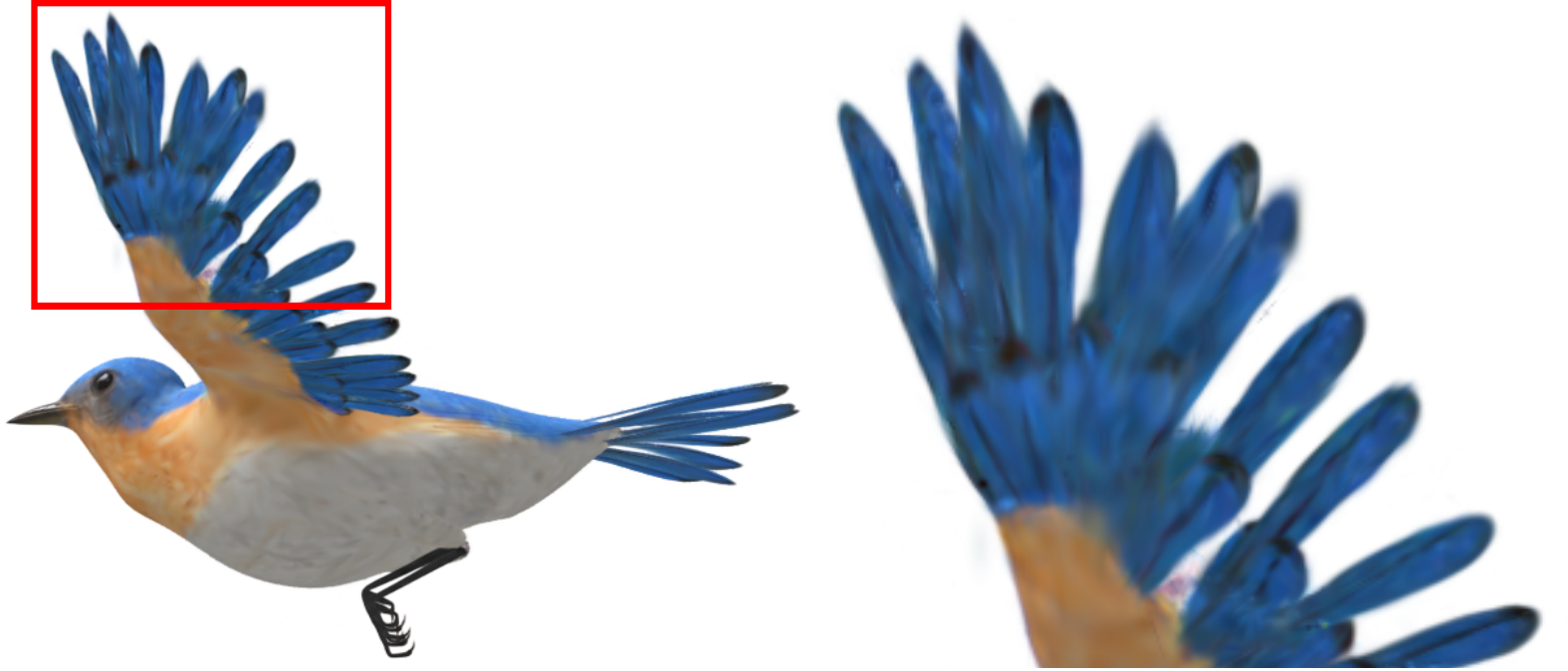}
    \caption{Full Model}
  \end{subfigure}
  \begin{subfigure}{0.42\linewidth}
    \centering
    \includegraphics[width=\linewidth]{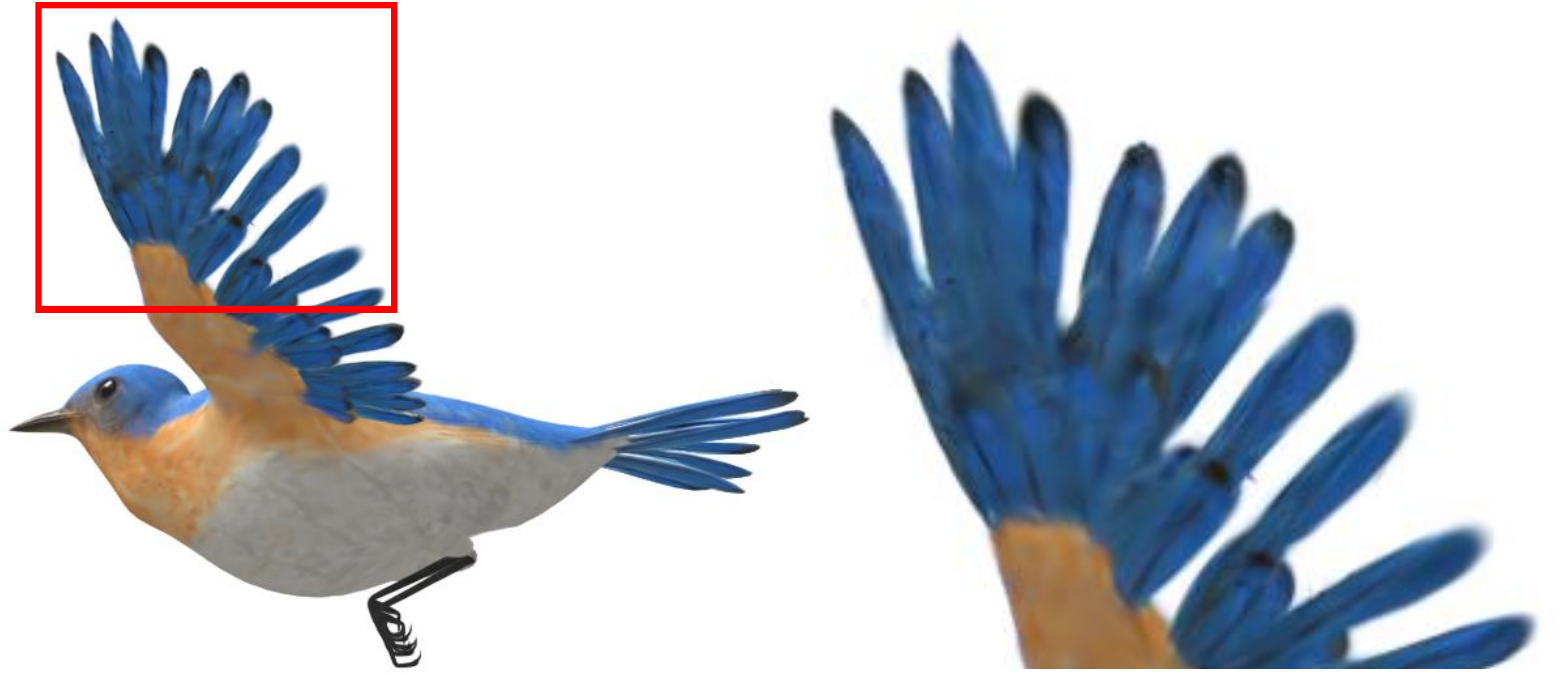}
    \caption{Single MLP}
  \end{subfigure}

  \vspace{2pt}

  \begin{subfigure}{0.42\linewidth}
    \centering
    \includegraphics[width=\linewidth]{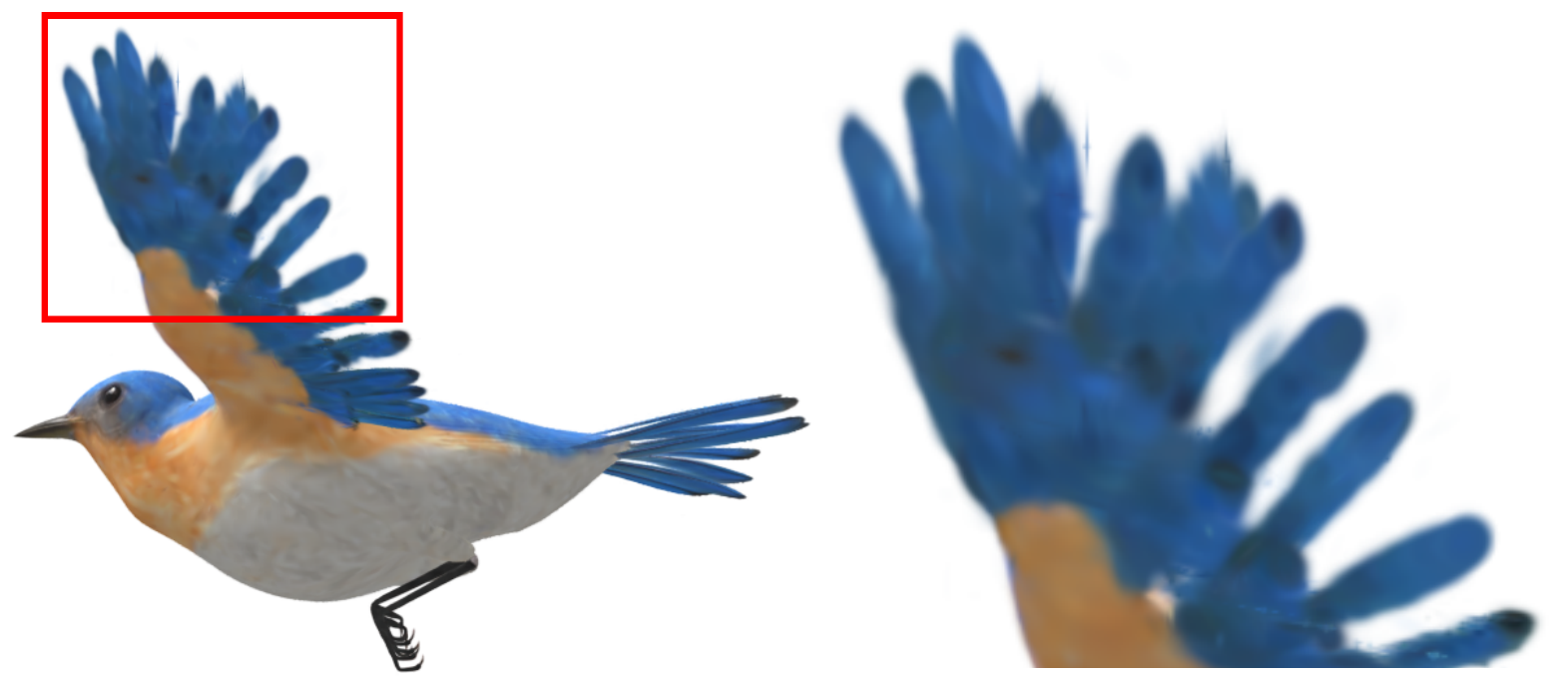}
    \caption{w/o Residual}
  \end{subfigure}
  \begin{subfigure}{0.42\linewidth}
    \centering
    \includegraphics[width=\linewidth]{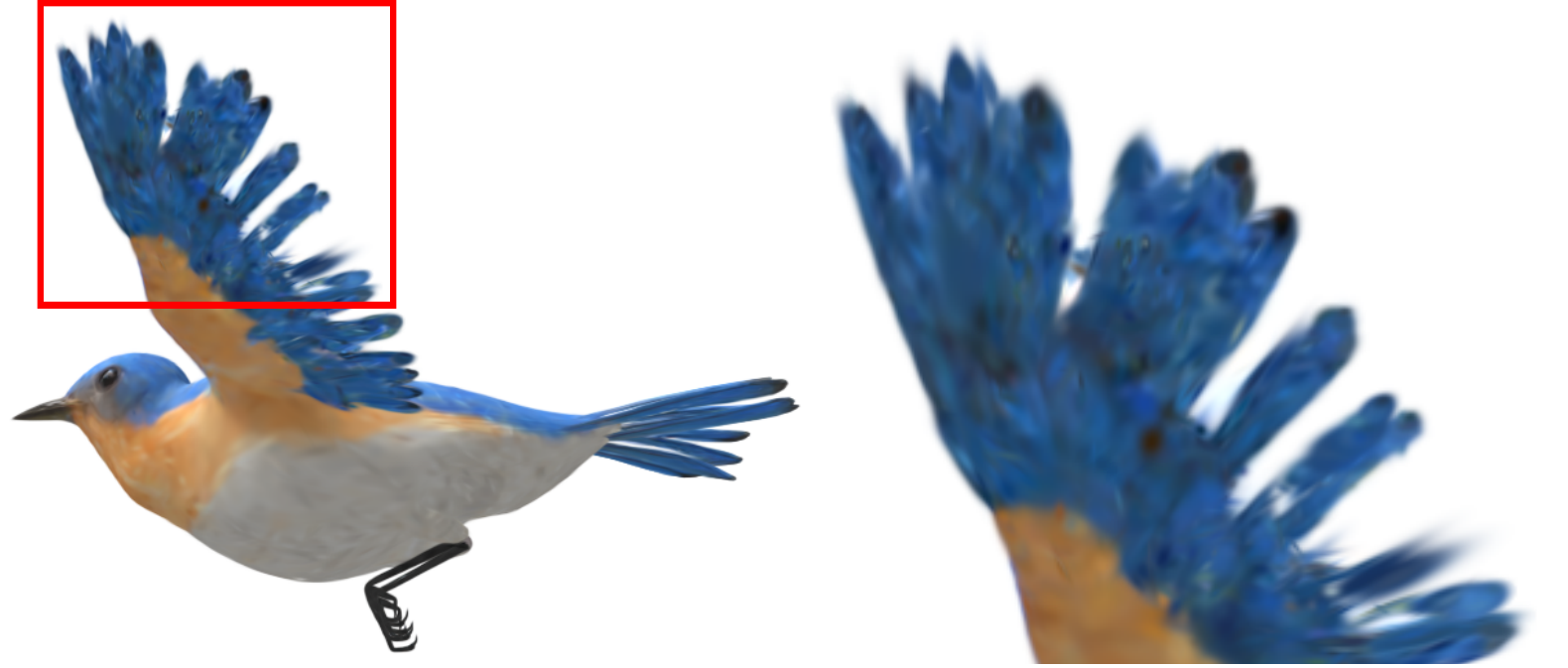}
    \caption{w/o Canonical}
  \end{subfigure}

  \caption{
  \textbf{Ablation: architecture components.}
  Removing or simplifying modules leads to degraded quality.}
  \label{fig:arch_ablation}
\end{figure}

\textbf{Architecture components.}
Table~\ref{tab:ablation} and Fig.~\ref{fig:arch_ablation} isolate each architectural module.
Collapsing Static MLP and DeformNet into a Single MLP~(b) ($-$0.78\,dB) removes the canonical-temporal separation, causing edge blurring.
Removing the residual branch (c) ($-$0.47\,dB) forces illumination variation into canonical geometry, producing blurring in fine-grained regions.
Removing the canonical branch (d) ($-$1.15\,dB) yields the largest drop, as the model loses its time-invariant reference and must reconstruct geometry and motion simultaneously.

\begin{table}[!h]
\centering
\caption{Ablation of architecture components on D-NeRF, showing their individual contributions to reconstruction quality under identical settings}
\label{tab:ablation}
\small
\begin{tabular*}{0.96\columnwidth}{@{\extracolsep{\fill}}l ccc}
\toprule
\textbf{Method} & PSNR$\uparrow$ & SSIM$\uparrow$ & LPIPS$\downarrow$ \\
\midrule
Full Model     & \textbf{36.982} & \textbf{0.9871} & \textbf{0.0136} \\
Single MLP     & 36.201 & 0.9836 & 0.0198 \\
w/o Residual   & 36.512 & 0.9854 & 0.0162 \\
w/o Canonical  & 35.834 & 0.9812 & 0.0221 \\
\bottomrule
\end{tabular*}
\end{table}

\section{Conclusion}
\label{sec:conclusion}

We presented GrainGS, a dynamic 3D Gaussian Splatting framework that
reduces gradient interference between canonical geometry and temporal
deformation. By combining static warm-up and stop-gradient isolation,
GrainGS preserves a more stable canonical reference, while
per-Gaussian deformation captures fine-grained local motion within a
compact anchor scaffold. The canonical-residual appearance
decomposition further provides explicit capacity for temporal
photometric variations, reducing their interference with geometric
motion modeling. Extensive experiments on D-NeRF and DG-Mesh
demonstrate that GrainGS achieves competitive SOTA
reconstruction quality while maintaining real-time rendering and
compact model storage.


\bibliographystyle{IEEEtran}
\bibliography{references}


\end{document}